\documentclass{article} % For LaTeX2e
\usepackage{colm2024_conference}

\usepackage{microtype}
\usepackage{hyperref}
\usepackage{url}
\usepackage{booktabs}

\usepackage{amsmath,amsfonts,bm}

\def\eqref#1{equation~\ref{#1}}
\def\Algref#1{Algorithm~\ref{#1}}

\def\1{\bm{1}}

\DeclareMathAlphabet{\mathsfit}{\encodingdefault}{\sfdefault}{m}{sl}
\SetMathAlphabet{\mathsfit}{bold}{\encodingdefault}{\sfdefault}{bx}{n}

\newcommand{\fig}[1]{Figure~\ref{#1}}

\newcommand{\tbl}[1]{Table~\ref{#1}}

\newcommand{\zc}[1]{\textcolor{red}{[Chao: #1]}}

\newcommand{\myparagraph}[1]{\vspace{-6pt}\paragraph{#1}}

\def\model{Learning Through Communication\xspace}
\def\modelshort{LTC\xspace}

\usepackage{mfirstuc,tabulary}

\usepackage{color,xcolor}
\usepackage{epsfig}
\usepackage{graphicx}

\usepackage{adjustbox}
\usepackage{array}
\usepackage{booktabs}
\usepackage{colortbl}
\usepackage{float,wrapfig}
\usepackage{hhline}
\usepackage{multirow}
\usepackage[toc,page]{appendix}

\usepackage{amsmath,amsfonts,amsthm,amssymb}
\usepackage{bm}
\usepackage{nicefrac}
\usepackage{microtype}

\usepackage{changepage}
\usepackage{extramarks}
\usepackage{fancyhdr}
\makeatletter \def\@fancyvbox#1#2{\vbox{#2}} \makeatother
\usepackage{lastpage}
\usepackage{setspace}
\usepackage{soul}
\usepackage{xspace}
\usepackage{indentfirst}

\usepackage{pifont}

\usepackage{algorithm, algorithmic}
\usepackage{listings}
\usepackage{enumerate}
\usepackage{mathtools}
\usepackage[symbol]{footmisc}

\usepackage{etoolbox}
\makeatletter
\AfterEndEnvironment{algorithm}{\let\@algcomment\relax}
\AtEndEnvironment{algorithm}{\kern2pt\hrule\relax\vskip3pt\@algcomment}
\let\@algcomment\relax
\newcommand\algcomment[1]{\def\@algcomment{\footnotesize#1}}
\renewcommand\fs@ruled{\def\@fs@cfont{\bfseries}\let\@fs@capt\floatc@ruled
  \def\@fs@pre{\hrule height.8pt depth0pt \kern2pt}%
  \def\@fs@post{}%
  \def\@fs@mid{\kern2pt\hrule\kern2pt}%
  \let\@fs@iftopcapt\iftrue}
\makeatother
\usepackage{arydshln}

\title{Adapting LLM Agents with Universal Feedback in Communication}

\author{\textbf{Kuan Wang}$^1$, \textbf{Yadong Lu} $^2$, \textbf{Michael Santacroce}$^{2}$, \textbf{Yeyun Gong}$^{3}$, \\ \textbf{Chao Zhang}$^{1}$, \textbf{Yelong Shen}$^{2}$ \\
$^1$Georgia Institute of Technology \ \ $^2$Microsoft Azure AI  \ \ $^3$Microsoft Research \\
\texttt{\{kuanwang, chaozhang\}@gatech.edu} \\
\texttt{\{yadonglu, misantac, yegong, yeshe\}@microsoft.com}
}

\colmfinalcopy % Uncomment for camera-ready version, but NOT for submission.
\begin{document}

\maketitle

\begin{abstract}

  % Recent advancements in large language models (LLMs) have shown potential for advanced agents.
  % To support online learning for these agents without extensive human supervision, we propose the Learning through Communication (LTC) paradigm, a novel training approach enabling LLM agents to enhance their skills through interactions with their environments and other agents.
  % Through iterative exploration and updating, LTC empowers the agent to assimilate short-term experiences into long-term memory.
  % To optimize agent interactions for task-specific learning, we introduce diverse communication patterns tailored for both single-agent and multi-agent environments.
  % We evaluated \modelshort on four datasets: ALFWorld (single-agent), HotpotQA (multi-agent collaboration), Chameleon (multi-agent competition), and GSM8k (multi-agent teacher-student).
  % On \textit{ALFWorld}, it exceeds the instruction tuning baseline by 12\% in success rate. On \textit{HotpotQA}, \modelshort surpasses the instruction tuned Llama-7B agent by 5\% in EM score, and it outperforms the instruction tuned 9x larger PaLM-62B agent by 0.6\%. On \textit{Chameleon}, the winning rate of LTC agent surpasses the baseline by 3.1\%. On \textit{GSM8k}, \modelshort outperforms the CoT-Tuning baseline by 3.6\% in accuracy. The results showcase the versatility and efficiency of the LTC approach across diverse domains. We will open-source our code to promote further development of the community.

  Recent advances in large language models (LLMs) have demonstrated potential for LLM agents.
To facilitate the training for these agents with both linguistic feedback and non-linguistic reward signals, we introduce Learning through Communication (LTC).
We design a universal buffer to store all the feedback, and an iterative pipeline to enable an LLM agent to explore and update its policy in an given environment.
To utilize our universal buffer for capturing agent interactions in various tasks, we introduce diverse communication patterns tailored for both single-agent and multi-agent environments.
We evaluate the effectiveness of our LTC approach on four diverse datasets: ALFWorld (single-agent), HotpotQA (multi-agent collaboration), Chameleon (multi-agent competition), and GSM8k (multi-agent teacher-student).
On these datasets, LTC outperforms supervised instruction fine-tuning baselines by 3.6\% to 12\%.
These results demonstrate the versatility and effectiveness of LTC in facilitating online adaptation for LLM agents.

  % Our universal buffer and pipeline are applicable to diverse agent settings, including single-agent and multi-agent environments. So we introduce 3 communication patterns tailored for sin
  % LTC is a unified paradigm that enables LLM agents to enhance their capabilities through interactions with their environments and other agents.
  % By engaging in iterative exploration and updating, LTC empowers the agent to integrate short-term experiences into long-term memory.

  % On ALFWorld, our LTC agent surpasses the instruction tuning baseline by an impressive 12\% in success rate.
  % On HotpotQA, LTC outperforms the instruction tuned Llama-7B agent by 5\% in EM score and even surpasses the instruction tuned 9x larger PaLM-62B agent by 0.6\%.
  % On Chameleon, the winning rate of our LTC agent exceeds the baseline by a significant margin of 3.1\%.
  % Lastly, on GSM8k, LTC outperforms the CoT-Tuning baseline by 3.6\% in accuracy.
  % These results underscore the versatility and efficiency of the LTC approach across diverse domains.

\end{abstract}

\section{Introduction}

Recent advances in large language models (LLMs)~\cite{ouyang2022training, bubeck2023sparks, wei2022emergent} have shed light on human-like LLM agents. In addition to design prompting methods~\cite{wei2022chain, yao2023react, wu2023autogen}, recent works also focus on how to train the LLMs agent use linguistic feedback and non-linguistic reward signals. The linguistic feedback is usually processed as instruction data to do Instruction Fine-tuning (IFT)~\cite{chung2022scaling, lee2023rlaif, honovich2022unnatural,wang2022self}, while the non-linguistic reward signals are generally used to do alignment with human preference~\cite{ouyang2022training, bai2022training, stiennon2020learning, leike2018scalable}.

While some scenarios provide agents with heterogeneous feedback,  existing methods can only utilize the feedback partially. For instance, in multiplayer board role-playing games, players generate a wealth of linguistic data, and the game concludes with definitive reward signals indicating victory or defeat. Current approaches employ the linguistic data for IFT~\cite{li2023camel, micheli2021language}, while the reward signals serve solely as a filtering criterion to select the ILF data instead of the objective of reinforcement learning.

% \begin{wrapfigure}{br}{0.45\textwidth}
% % \begin{table}[ht]
% % \vspace{-10pt}
% \centering
% \small

% % \begin{figure}[h!]
%   \centering
%   \includegraphics[scale=0.29]{figures/teaser.png}
%   \caption{Our LTC framework could continually learn from the collected iterative interaction trails with task-specific fine-tuning, and there is no need for few-shot examples as the prefix of the prompts in the inference time, which significantly reduces the token length of the agent inputs and reduces the memory and computation cost \zc{ This figure is too tall, maybe change the layout to make it horizontal and span the entire column.} .}
%   \label{fig:teaser}
% % \end{figure}

% \end{wrapfigure}

\begin{figure*}[t]
\vspace{-30pt}
  \centering
  \includegraphics[scale=0.225]{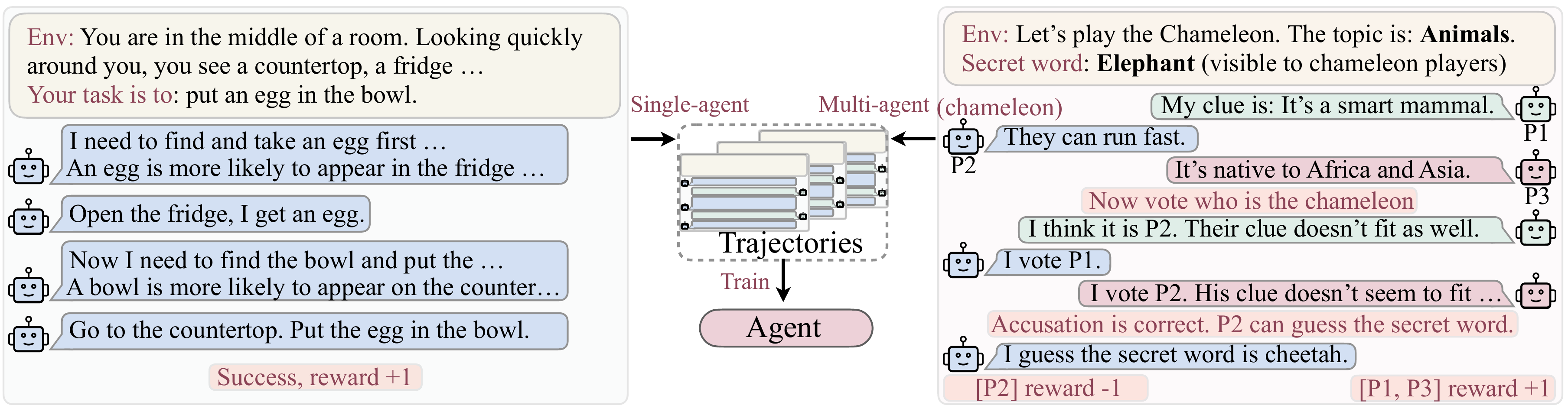}
  \vspace{-10pt}
  \caption{
    The \modelshort framework is adept for both single-agent and multi-agent environments. Within these environments, agents have the capability to persistently engage in exploration and interaction to collect trajectories through various communication patterns. Concurrently, \modelshort facilitates the training of these agents utilizing the data acquired from their exploratory activities. This process enables the agents to autonomously adapt to their respective environments, negating the necessity for human supervision.
    % removing the need for long and handcrafted few-shot examples as the prefix of the prompts in the inference time, providing a trade-off option between training efficiency and the inference cost.
    % which significantly reduces the token length of the agent inputs and reduces the memory and computation cost. 
    % \zc{This is such a long sentence. I did some editing for you.  The purpose of this figure is to help readers understand the high-level idea of your master. Many things mentioned in the original sentence, such as the benefits, are not really necessary. ``In \modelshort, the LLM agent continuously interact with the environment and other agents to collect communication trajectories. These trajectories capture the agent's states, actions and rewards. They are then used to fine-tune the LLM agent via policy gradient methods, enabling the agent to adapt to the environment without requiring human annotation efforts''.}
    }
  \label{fig:teaser}

% \vspace{-15pt}
\end{figure*}

To address this gap, we propose a universal framework, named Learning through Communication (LTC), to train LLM agents with both linguistic feedback and non-linguistic reward signals.
We design a universal buffer to store all the feedback, and an iterative pipeline to enable an LLM agent to explore and update its policy in an given environment.
Each iteration of LTC comprises two distinct phases: (1) \emph{\textbf{Exploration}}: During this phase, the agent interacts with the environments and other agents to gather diverse trajectories (linguistic) and reward signals (non-linguistic) into the universal buffer. (2) \emph{\textbf{Updating}}: In this phase, the agent's model is updated based on the collected data in the universal buffer.
For updating, LTC combines the language modeling loss and the PPO loss to strike a balance between language consistency and reward signals As the pivot of the iterative pipeline, the replay buffer is updated after each exploration phase, and a subset of the buffer is sampled for the updating phase.

% to engage in online learning across various environments and tasks through an iterative pipeline.
% Our iterative process updating phase, we leverage reinforcement learning (RL). 
% Furthermore, we introduce a unified replay buffer to store trajectories from diverse environments. Each generated token is treated as an action within the RL formulation. Distinct masks are applied to tokens generated by the system, LTC agents, and other agents, allowing for applying different loss functions to different token types.

% Through this iterative pipeline, the agent continuously engages in online learning in new environments.

To  universally supports  linguistic feedback and non-linguistic reward signals during communication, we design the replay buffer structure as
a trajectory  of tokens sequences (\fig{fig:buffer}). Such a replay buffer structure  is applicable to diverse tasks, including single-agent and multi-agent environments. To facilitate collecting trajectories with linguistic data and reward signals, we devised three communication patterns:
(1) \emph{\textbf{Single-agent Monologue}}: This pattern allows a single agent to collect trajectories contain linguistic data and receive reward signals from the environments.
(2) \emph{\textbf{Multi-agent Dialogue}}: This pattern enables multiple agents to interact with each other and external tools to collect linguistic data, and utilize reward signals provided by the environments.
(3) \emph{\textbf{Teacher-student Dialogue}}: This variant of multi-agent dialogue that collect the linguistic feedback and non-linguistic reward signals provided by a teacher agent instead of the environment.

% involves a teacher agent providing linguistic feedback and non-linguistic reward signals instead of the envirm.

We evaluate LTC method on several representative datasets: \textit{ALFWorld} for decision-making, \textit{HotpotQA} for knowledge-intensive reasoning, and \textit{GSM8k} for numerical reasoning.
Throughout these experiments, \modelshort consistently outperforms the baselines.
In \textit{ALFWorld}, \modelshort outperforms the strong instruction tuning baseline by 12\% on success rate, even in the challenging Pick 2 task.
This shows that our communication mechanism enables the agent to learn from its experiences for task solving.
On \textit{HotpotQA}, \modelshort outperforms the instruction tuning baseline by 5\% on EM score, and our Llama-7B based agent even obtains slightly better (0.6\%) performance than the ReAct-Tuning baseline which uses 9$\times$ larger PaLM-62B model.
On \textit{GSM8k}, \modelshort also beats the CoT-Tuning baseline by 3.6\% on accuracy.
These results highlight the adaptability and effectiveness of LTC approach across varied domains.

Our key contributions are:

\begin{enumerate}
  \item \textbf{Learning through Communication (LTC)}:
  We propose a universal framework, named Learning through Communication (LTC), to train LLM agents with \textbf{both linguistic feedback and non-linguistic reward signals}. We design a universal buffer to store all the feedback, and an iterative pipeline to enable an LLM agent to explore and update its policy in an given environment.
        % We introduce a universal framework called Learning through Communication (LTC), which enables the LLM agent to autonomously adapt to novel environments and tasks through an iterative pipeline.

  \item \textbf{Task-specific Communication Patterns}:
        The LTC paradigm allows for flexible design of communication patterns tailored to different tasks. We introduce three specific patterns: Single-agent Monologue, Multi-agent Dialogue, and Teacher-student Dialogue. These patterns can be combined to generate diverse structured interactions and feedback signals for agent training, catering to various task types.

  \item \textbf{Empirical Study and Findings}:
        We conduct rigorous experiments on public benchmark tasks to demonstrate the effectiveness of LTC. Our results indicate that LTC can be a superior approach compared to instruction-tuning or prompting baselines.

\end{enumerate}

\section{Related Work}

\subsection{Instruction Tuning}

\begin{figure*}[t]
  \centering
  \vspace{-40pt}
  \includegraphics[scale=0.28]{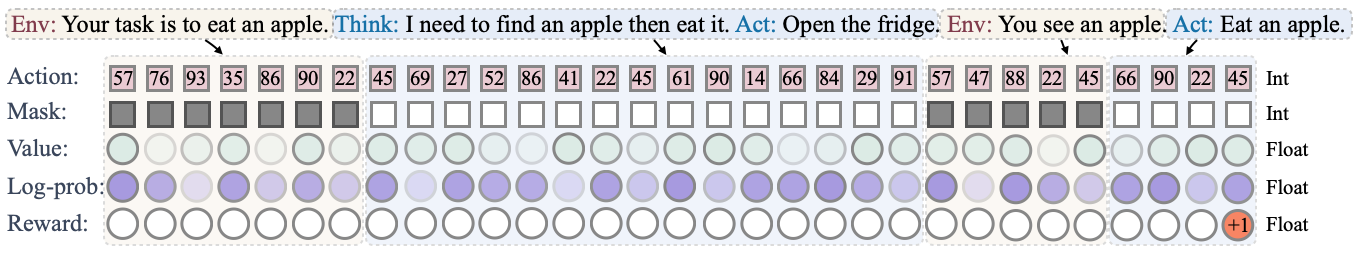}
  \vspace{-10pt}
  \caption{The buffer data is a serial of integer/float sequences. We treat each token id as the action in our reinforcement learning formula. We also save its corresponding mask indicating the source of the token, the value from the critic model, the log-prob indicating the log-likelihood when sampling the action and the reward from the environment/other agents.}
  \label{fig:buffer}
  % \vspace{-15pt}
\end{figure*}

Instruction tuning (IT) is an important  technique for improving the capabilities and controllability of LLMs~\cite{radford2019language, brown2020language, wei2022emergent, qin2023chatgpt, openai2023gpt4, chowdhery2022palm, touvron2023llama}. Many studies have been dedicated to instruction data generation and selection~\cite{chung2022scaling, wang2022self, lee2023rlaif}. For instance, Unnatural Instructions~\cite{honovich2022unnatural} is created by using the Super-Natural Instructions dataset~\cite{wang2022super} as a seed to prompt InstructGPT~\cite{ouyang2022training}. Self-Instruct~\cite{wang2022self} employs a recursive pipeline that generates instruction data from hand-crafted seed tasks using ChatGPT~\cite{chatgpt}.
Other studies focus on fine-tuning pre-trained LLMs with instruction data. BLOOMZ~\cite{muennighoff2022crosslingual} is initialized with BLOOM~\cite{scao2022bloom} and then fine-tuned using the xP3 instruction dataset~\cite{muennighoff2022crosslingual}. Flan-T5 is initialized with T5~\cite{2020t5} and fine-tuned with the FLAN dataset~\cite{longpre2023flan}. Additionally, after the release of LLaMA~\cite{touvron2023llama}, many works have utilized it as the base model for instruction tuning, such as Alpaca~\cite{taori2023alpaca}, Vicuna~\cite{chiang2023vicuna}, and GPT-4-LLM~\cite{peng2023instruction}.
Some papers explore alignment fine-tuning using RLHF~\cite{ouyang2022training, bai2022training, stiennon2020learning, leike2018scalable}. InstructGPT~\cite{ouyang2022training} employs GPT-3 for supervised fine-tuning  on a human-filtered instruction dataset, followed by training a reward model and using PPO~\cite{schulman2017proximal} for RLHF. Claude investigates RLHF ~\cite{bai2022training} and constitutional approaches \cite{bai2022constitutional} for making LLMs both harmless and helpful. DPO~\cite{rafailov2023direct} fine-tunes the LLMs to align with human preferences by directly optimizing a classification problem on preference data instead of RLHF. While these prominent research works focus on aligning LLMs for general instruction-following, our objective is to adapt LLM agents for specific tasks or environments.

\subsection{LLM Agents}
% \myparagraph{LLMs as Agents}

LLMs have demonstrated the potential to act as advanced agents~\cite{ouyang2022training, bubeck2023sparks, wei2022emergent}, and significant progress has been made in developing versatile LLM agents~\cite{weng2023prompt, sumers2023cognitive, Park2023GenerativeAgents, liu2023training, Lin2023SwiftSageAG, xu2023rewoo} and benchmarks ~\cite{scienceworld2022, deng2023mind2web, liu2023agentbench}.
For planning, Chain-of-Thought (CoT\cite{wei2022chain}) prompts the model to think step by step, by decomposing complex tasks into smaller and simpler steps.
Self Consistency~\citep{wang2022self-consistency,wang2022rationale} extends CoT by using ensembles of predictions to improve consistency of the LLM.
Inner Monologue \cite{huang2022inner} leverages environment feedback to enhance LLMs' planning and processing capabilities in embodied robotics tasks without extra training.
ReAct~\cite{yao2023react} integrates reasoning and action taking, expanding the action space to include both task-specific discrete actions and language.
Reflexion~\cite{shinn2023reflexion} equips agents with dynamic memory and self-reflection capabilities to improve reasoning by using continuous trials in the same environment as feedback.
Recent research has also shown that LLMs can be augmented as an autonomous agent to use \emph{external tools} to solve problems in interactive environments.
These techniques include retrieval augmentation \cite{shi2023replug, yao2023react, izacard2022few}, math tools \cite{schick2023toolformer, yao2023react, lu2023chameleon}, and code interpreters \cite{gao2022pal, wang2022code4struct}.
Prior works also have explored using multiple LLMs in a collaborative setting to solve complex tasks~\cite{hong2023metagpt, qian2023communicative, li2023camel, wang2023unleashing, talebirad2023multi, akata2023playing}.
Open-source projects like AutoGPT~\cite{Gravitas2023}, GPT-Engineer~\cite{AntonOsika2023}, and BabyAGI~\cite{yoheinakajima2023} also showcase the potential of LLM not just in generating content but also as a general problem solver.
Most of the above methods are based on either human-designed few-shot prompting examples, or finetuning with pre-collected instruction datasets. 
Our \modelshort is not a few-shot prompting method and we focus on adapting the agent by collecting training data automatically by exploration.

\section{\model}

% \subsection{Iterative Pipeline}

\begin{figure*}[t]
\vspace{-40pt}
  \centering
  \includegraphics[scale=0.23]{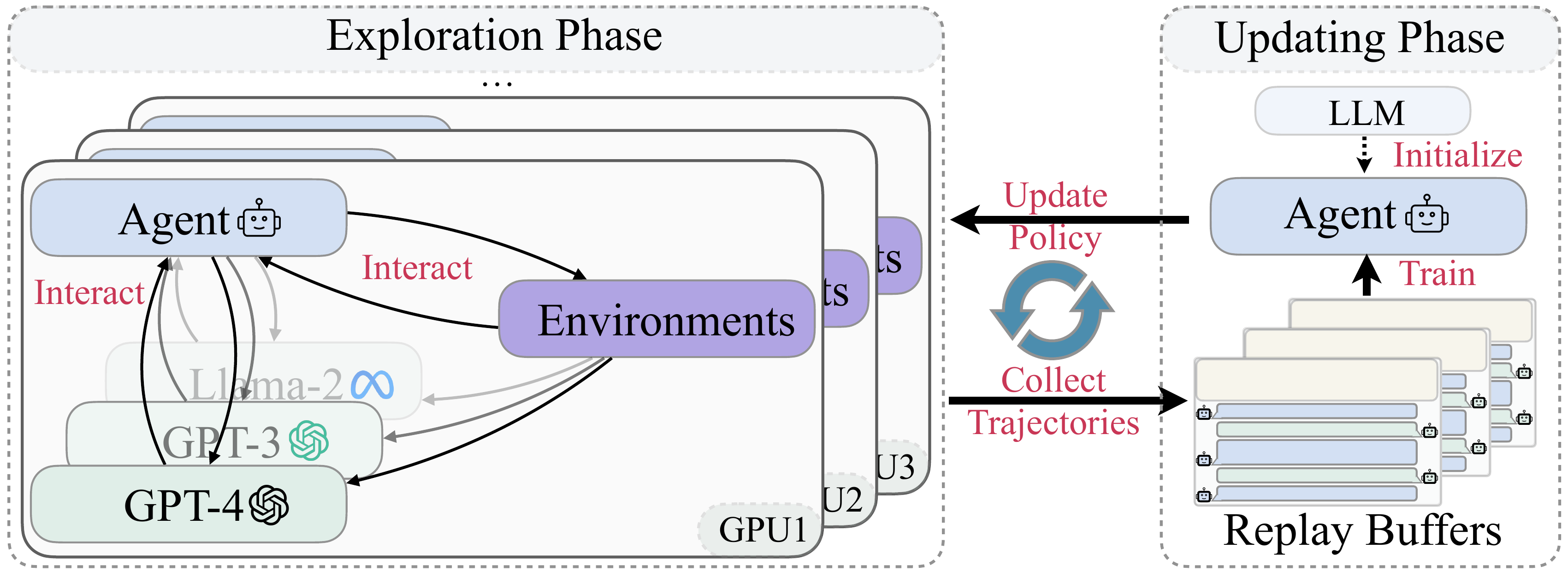}
  \vspace{-10pt}
  \caption{LTC has an iterative two-phase framework. During the exploration phase, the agent proactively explores new environments and communicates with other agents, gathering the trajectories to update the replay buffer. Then the agent is trained for updating the policy in the updating phase. }
  \label{fig:overview}
  % \vspace{-10pt}
\end{figure*}

We design  \model (\modelshort), an iterative training method for LLM agents to continuously adapt to new environments. As shown in Figure~\ref{fig:overview}, LTC iterates between two phases: (1) An exploration phase where agents can interact with new environments and other agents to collect trial data with feedback, and (2) a updating phrase to fine-tune the agent to update the policy.

% \subsection{Communication Session}

\subsection{Exploration Phase}
At the start of each iteration, the agent explores the environments to get the trajectories and the reward signal data. We denote these data as a tuple: $\mathcal{S}=(\mathcal{T}, \mathcal{M}, \mathcal{R})$, where \( \mathcal{T} = \{t_1, t_2, \dots, t_n\} \) represents the text data generated by the communication process during agent exploration, \( \mathcal{M} = \{m_1, m_2, \dots, m_n\} \) with \( m_i \in \{0, 1, 2\} \) indicates the source of the text data (system or agents), \( \mathcal{R} = \{r_1, r_2, \dots, r_n\} \) with \( r_i \in \{-1, 0, 1\} \) represents the reward signals provided by either the system or the agents. We demonstrate the the details of this data structure in \fig{fig:buffer}, $\mathcal{M}$ is the mask list, and $\mathcal{R}$ is the reward list. In PPO training, both the value list and the log-prob list correspond directly to the action list. For brevity, we denote these three lists together as $\mathcal{T}$ here. Please see Appendix~\ref{sec:appendix_b} for more detaills.

\begin{figure*}[t]
\vspace{-40pt}
  \centering
  \includegraphics[scale=0.26]{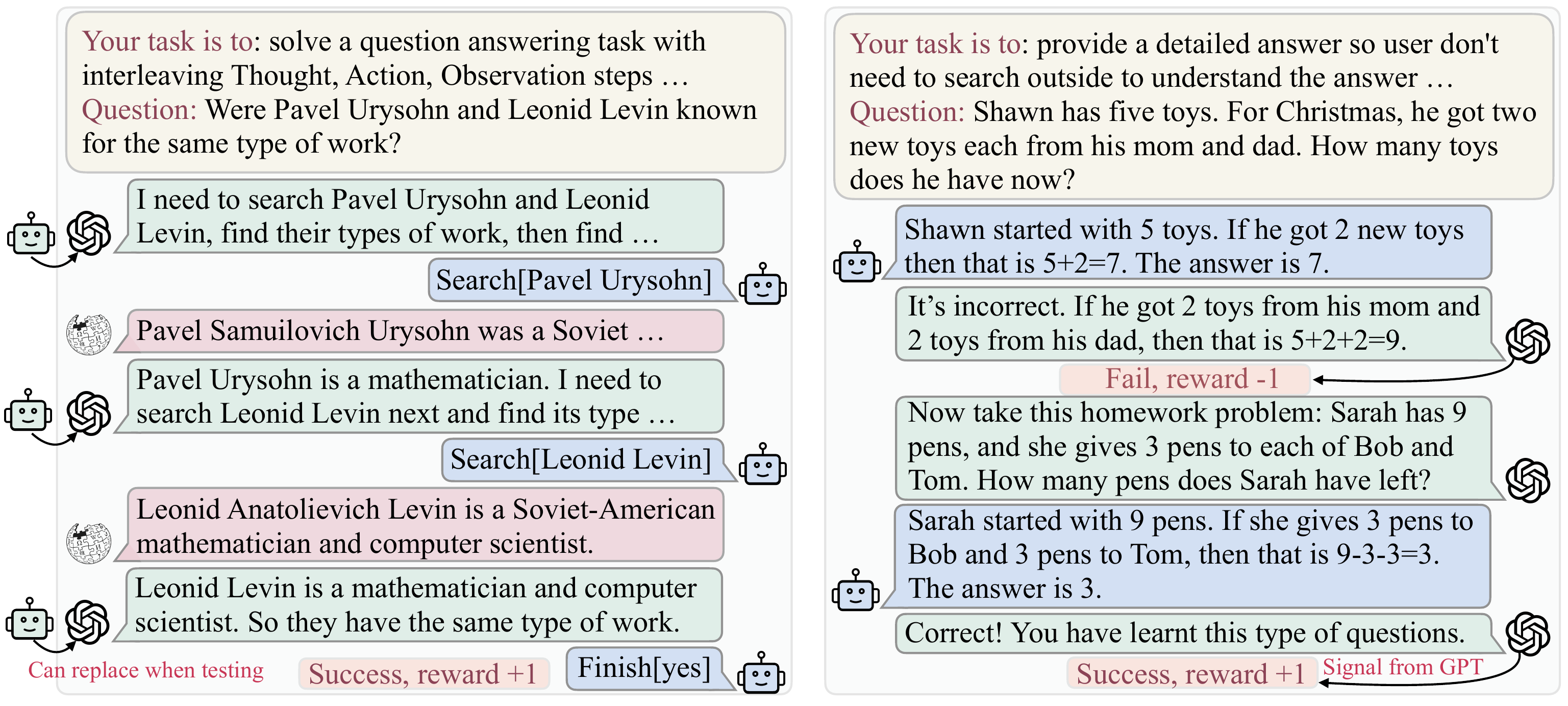}
  \vspace{-10pt}
  \caption{The toy examples to demonstrate communication patterns: 1) the left figure is the Multi-agent Dialogue pattern, where two agent play different roles to collaborate on the task. The thinker agent is responsible for analyzing the situation and give suggestion to the actor agent who is responsible for making decisions. We can just assign the LTC agent to play the thinker agent when testing without GPT-4 agent. 2) the right figure is the Teacher-student Dialogue pattern, where the student agent starts with an initial answer to the current question, and then the teacher directly corrects the answer with a reward. To help the student improve ability instead of just memorizing the solution, the teacher will generate another analogous question to ask the student. Eventually, the student gives a new answer for this analogous question and gets a new reward signal from the teacher.}
  \label{fig:patterns}
  % \vspace{-10pt}
\end{figure*}

To collect the trajectories data $\mathcal{S}=(\mathcal{T}, \mathcal{M}, \mathcal{R})$ from different types of tasks, we design the communication patterns for these tasks. Here we provide three communication patterns:

\begin{itemize}
   % Motivated by Inner Monologue~\cite{huang2022inner},
  \item \textbf{Single-agent Monologue}: Single-agent Monologue is a single-agent soliloquy style communication pattern, designed for general instruction following tasks (\Algref{alg:monologue}). It split the tasks into step by step like ReAct and CoT, and their own trajectories with system rewards are collected to train themselves at the same time with their exploration.
  % The agents need to decompose the hard task into smaller and simpler steps~\cite{wei2022chain}, and \textbf{the same agent will play different roles} in this steps to solve the problem. 
  % \zc{ It's such a long sentence, break it down. Focus on describing the key idea. Do you need to really mention it's inspired by inner monologue? That does not help understanding. Just talk about what this monologue pattern is.  Can you give one sentence and precise definition for monologue?}  
   \fig{fig:teaser} left is a toy example of \textit{ALFWorld} to demonstrate the Monologue pattern with a single agent. This agent soliloquizes to think the situation and take the actions to explore the environment and finally get the reward provided by the environment. This pattern is based on the think and act steps in the ReAct formulation~\cite{yao2023react}, we design the training buffer collection process to make it aligh with our reinforcement learning formulation.
        % \zc{If I remember correctly, React has also done fine-tuning using the traces. In that case, how do we differentiate from it?  Is it that  we have both positive and negative rewards, so we can use the PPO training and better use the experience?  Reviewers may also have the same doubts, as fine-tuning LLM  agents from experiences has been explored in prior works. We need to emphasize the difference in the introduction and related work.}
  \item \textbf{Multi-agent Dialogue}:  Multi-agent Dialogue is a multi-agent discussion style pattern (\Algref{alg:dialogue}). 
  % \zc{ Again, it will be best if you can give one sentence, clear definition of the dialogue pattern first, and then elaborate.  Don't start describing by saying it is for xxx, it supports xxx, but give the important definition first.} 
  It is designed for multi-agent collaborating and competing tasks, where multiple agents will play their role by speaking or taking actions in a certain order and \textbf{a final reward will be given by the environment} based on the performance of the agents. The left figure of \fig{fig:patterns} is a toy example of \textit{HotpotQA} to illustrate this pattern for collaborating, where the GPT-4 agent play as a thinker to analyze the situation and give suggestions to the actor agent who is responsible for making decisions. The reward in \textit{HotpotQA} is the correctness of the answer obtained by two agents. And we can use their communication data to train the LTC agents do both thinker and actor so that they can learn how to cooperate with each other to solve the task. The right figure of \fig{fig:teaser} is a toy example of Multi-agent Dialogue for a competing game task Chameleon, where three agents play different roles. The reward is the win or loss of the game, so they need do with deduction and bluffing in the communication process to win the game. And their games trajectories will be use in LTC iterations to boost the agents.

  % where multiple knowledge sources are needed for a comprehensive answer. 
  % Similar to the Monologue, the Dialogue pattern also has multiple roles, but \textbf{the roles are played by different agents}. This pattern also supports communicating with external tools and other agents, which can be easily achieved by changing the roles' prompts or replacing the agents behind the roles. The left figure of \fig{fig:patterns} is a toy example of \textit{HotpotQA} to illustrate this pattern, where the GPT-4 agent play as a teacher to analyze the situation and give suggestions to the student agent who is responsible for making decisions. We can just assign the student agent to play as the teacher agent when testing for environments without GPT-4. So we can use this pattern to collect the buffer with GPT-4 as teacher agent and train the student to play as a teacher to improve the ability of the student agent to solve problems independently.

  \item \textbf{Teacher-student Dialogue}: Teacher-student Dialogue is a teacher-student style pattern for powerful agents to teach the novice agents (\Algref{alg:analogue}). We design this pattern for complex analytical tasks such as numerical reasoning, which require extensive analytical examples for agents to improve the specific reasoning ability lacking in the pretrained models.  Teacher-student Dialogue pattern has two roles (student and teacher) played by two agents, however, in addition to the linguistic feedback, the teacher roles can \textbf{directly provide the non-linguistic reward signals}, which are all provided by the system (environments) in the previous pattern. The right figure of \fig{fig:patterns} is a toy example with GSM8k to demonstrate how the student agent communicates with the teacher agent in a homework-correcting style. In the math question environment, the student agent starts with an initial answer to the current question, then the teacher directly corrects the answer with a reward. To help the student improve ability instead of just memorizing the solution, the teacher will generate another individual question and provide a new reward to the student. 
  % Eventually, the student gives a new answer for this analogous question and gets a new reward signal from the teacher.

\end{itemize}

\subsection{Updating phase}

In the updating phase, the LLM agent model could be optimized through the conversation sessions collected in the exploration stage. Given a example session $\mathcal{S}=(\mathcal{T}, \mathcal{M}, \mathcal{R})$, we mainly utilize two training objects for model training.

\begin{itemize}
        \item Language model Objective: $\mathcal{L}_{\text{LM}}$ encourages the model to learn from the trajectory $\mathcal{T}$, serving as an unsupervised learning schema to help model for behavior cloning from other agents' response or predicting system feedbacks.
        %is defined as the cross entropy between the agent and its generations which have a positive reward, akin to self-improving model schemes \cite{huang2022large, rafailov2023direct}. By training on these generations, the agent is further encouraged to give generations which yield positive rewards.
%The term "Reinforcement Objective" typically refers to a goal or target in reinforcement learning, where an agent learns to make decisions to maximize a reward signal. In this context, the reinforcement objective guides the agent's behavior and learning process. If you have a specific context or question related to the reinforcement objective, please provide more details, and I'll be happy to assist further.

        \item Reinforcement Objective: $\mathcal{L}_{\text{reinforce}}$ optimizes the model by maximizing the expectation reward provided by environment or a teacher agent (i.e., GPT-4 \cite{openai2023gpt4}). It is an goal-oriented objective, and allows the model to learn through both positive and negative signals in the communication session.
\end{itemize}

Thus, the overall training objective for LTC combines the above two terms:
\begin{equation}
    \mathcal{L}_{\text{LTC}}(\mathcal{S}) = \beta \mathcal{L}_{\text{LM}} (\mathcal{T}) + \mathcal{L}_{\text{reinforce}}(\mathcal{S}), 
\end{equation}
where $\beta$ is a balancing hyper-parameter. The off-policy  PPO algorithm \cite{schulman2017proximal} is utilized for optimizing $\mathcal{L}_{\text{reinforce}}(\mathcal{S})$, and it can be further breakdown into policy loss, value loss and policy entropy regularization terms in implementation. The vanilla PPO algorithm takes the triplet $(\text{state}, \text{action}, \text{rewards})$ for training. In this case, we sample from the trajectories $(\mathcal{T}_{<i}, t_i)$ for simulating the state-action pairs, specifically, we only keep the tokens generated by agent model itself as actions for policy updating. 

% https://di-engine-docs.readthedocs.io/en/latest/12_policies/ppo.html

\section{Experiments}
\label{sec:exp}
% \kuan{emphsis we do not use GPT-4 in the testing process.}

\subsection{Datasets}
% \myparagraph{Datasets}
% \zc{ If you run out of space, you can move the experiment setup details and dataset details to the appendix.}

% We conduct extensive experiments on \textit{ALFWorld}~\citep{shridhar2020alfworld}, \textit{HotpotQA}~\citep{yang2018hotpotqa} and \textit{GSM8k}~\citep{cobbe2021gsm8k}. Each of these datasets exemplifies a different type of task, representing decision-making, knowledge-intensive reasoning, and numerical reasoning, respectively.

We conducted experiments on four datasets: \textit{ALFWorld}~\citep{shridhar2020alfworld}, \textit{HotpotQA}~\citep{yang2018hotpotqa}, Chameleon \cite{ChatArena} and \textit{GSM8k}~\citep{cobbe2021gsm8k}.
Each of these datasets represents a different environment type, namely single-agent, multi-agent collaborating, multi-agent competing, and teacher-student, respectively. And different communication patterns are used: Single-agent Monologue for \textit{ALFWorld}, Multi-agent Dialogue for \textit{HotpotQA} and Chameleon \cite{ChatArena}, and Teacher-student Dialogue for \textit{GSM8k}.

\myparagraph{\textit{ALFWorld}}
\textit{ALFWorld} (Figure~\ref{fig:teaser}) is a text-based game that follows the ALFRED benchmark~\citep{shridhar2020alfred}. In this game, agents are presented with six types of tasks that involve navigating a simulated household environment using textual actions. With over 50 locations to explore, these tasks demand strategic planning and thorough exploration. Following ~\citep{shridhar2020alfworld}, we utilize the train set that  consists of 3553 environments for training our model and the baselines; and we use the unseen test set that  comprises 134 environments for evaluatation.

\myparagraph{\textit{HotpotQA}}

\textit{HotpotQA} is a question-answering dataset that focuses on multi-hop reasoning based supporting facts, with the goal of improving the explainability of QA systems. In this dataset, agents are required to reason across two or more Wikipedia passages to derive answers. We initialize the environments using only the text of the questions, meaning that agents are provided with the question and task description but do not have access to supporting paragraphs. To support their reasoning, agents must either rely on their internal knowledge or interact with an external Wikipedia tool to retrieve the necessary information. 
% The action space includes: 1) \textbf{search[entity]} for obtaining initial content or similar suggestions, 2) \textbf{lookup[string]} for retrieving relevant sentences, and 3) \textbf{finish[answer]} to conclude tasks. 
For training, we sample the environments from the training set, which consists of 90,447 QA-pairs. For evaluation, we run 500 random examples from the test set, following~\citep{yao2023react}.

\myparagraph{\textit{Chameleon}}
Chameleon is a multi-player social deduction game environment implemented by the ChatArena \cite{ChatArena}. There are two roles in the game, chameleon and non-chameleon. The topic of the secret word will be first revealed to all the players. Then the secret word will be revealed to non-chameleons. Non-chameleons try to identify the chameleon without giving away the secret word, while the chameleon tries to blend in and guess the word. The game involves giving clues, voting on who the chameleon might be, and a final guess from the accused chameleon. We use [3, 4, 5] players setting to train and test the agents' performance.

\myparagraph{\textit{GSM8k}}
The \textit{GSM8k} dataset is a collection of 8.5K math problems for grade school students. These problems have been crafted by human experts to ensure linguistic diversity. The dataset is divided into two sets: 7.5K problems for training and 1K problems for testing. Each problem in the dataset requires 2 to 8 steps of reasoning to arrive at the solution. The problems primarily focus on fundamental arithmetic operations like addition, subtraction, multiplication, and division.

% \textit{GSM8k} is a curated dataset consisting of 8.5K linguistically diverse grade school math word problems, handcrafted by human experts. Divided into 7.5K training and 1K test problems, these problems demand 2 to 8 steps of reasoning, primarily requiring basic arithmetic operations such as addition, subtraction, multiplication, and division to deduce the answer.
% \vspace{-10pt}
\subsection{Settings}
% \subsection{Implementation and training details}
% \myparagraph{Implementation and training details.}

\myparagraph{Model Architecture}
% We use the Llama~\cite{touvron2023llama} as our base model, but slightly modified its architecture for our RL setting. To generate state values corresponding to the action tokens, we introduce an additional linear layer at the bottom of the LlamaDecoderLayer, which serves as the value head.
We use a modified version of Llama~\cite{touvron2023llama} as the base model. To generate state values corresponding to the action tokens, we introduce an additional linear layer to serve ast the value head. 
% \michael{We use a modified version of LLaMA~\cite{touvron2023llama} as the base model. To generate state values corresponding to the action tokens, we introduce an additional linear layer to serve ast the value head. }
This value head acts as an auxiliary output module, and the output values are processed using the $tanh()$ function to ensure they fall within the range of (-1, 1).
This adaptation for RL has also been discussed in prior studies \cite{santacroce2023efficient}.
% \kuan{has personal information issue? cite or not? }

\myparagraph{Agent Pre-training}
\label{par:pretraining}
We use the Llama-7B model~\citep{touvron2023llama} for our LLM agent. To enhance the agent's ability to follow task-specific instructions, we initialize it by instruction fine-tuning (IT). And this initialized agent works as the baseline for a fair comparison.  This step is crucial because the original Llama-7B model, without prior instruction fine-tuning, struggled to follow task instructions and generation sensible actions in the environments. 
% \zc{Is this initialization also performed for all the bass lines? It is important to conduct this initialization for a fair comparison. If you have indeed performed this initialization, please make sure to mention it.} 
To collect data for instruction fine-tuning, we employ GPT3/4 as our agent to explore the environments created from the training set. We then filter out negative examples and retain positive examples to train the initial agent. For both the \textit{ALFWorld} and \textit{HotpotQA} datasets, we leverage GPT3 (specifically, text-davinci-003). However, for the \textit{GSM8k} dataset, we use GPT4 due to GPT3's inadequate performance in handling mathematical problems, which resulted in a scarcity of positive examples.

\myparagraph{Training details}
% \myparagraph{Distributed Training}

% We use AdamW~\citep{loshchilov2017decoupled} as the optimizer and set the batch size to 32. The learning rate is 2e-4. The sampling environment sizes designated for agents to explore and gather new training buffers are set at 256 for \textit{ALFWorld}, 512 for \textit{GSM8k}, and 1024 for \textit{HotpotQA} in each iteration. We use LoRA~\cite{hu2021lora} to do Parameter-Efficient Fine-Tuning (PEFT), and we set R to 16, ALPHA to 16. We use 4 nodes of 8$\times$A100 GPUs for the distributed training and exploration of  \textit{HotpotQA} and \textit{GSM8k} experiments, while we use only 1 node of 2$\times$A100 GPUs for each experiment of \textit{ALFWorld}.

We utilize the AdamW optimizer~\citep{loshchilov2017decoupled} with a batch size of 32. The learning rate is set to 2e-4. In each iteration, the sizes of new environments for agents to explore are: 256 for \textit{ALFWorld}, 512 for \textit{GSM8k}, and 1024 for \textit{HotpotQA}.
For parameter-efficient fine-tuning, we employ LoRA~\cite{hu2021lora} with hyperparameters $R=16$ and $\alpha=16$. For distributed training, we utilize 4 nodes with 8$\times$A100 GPUs on \textit{HotpotQA} and \textit{GSM8k}. For the experiments on \textit{ALFWorld}, we use 1 node with 2$\times$A100 GPUs due to the dataset's small scale.

% \zc{why?  Add a short explanation. } .
% \kuan{since I only have 2GPUs now}

\myparagraph{Baselines}

% \textbf{ReAct}~\citep{yao2023react} uses a subset of training cases to create ReAct-formatted trajectories with sequences of thought-action-observation as prompts for diverse tasks. For knowledge-intensive reasoning tasks like \textit{HotpotQA}, ReAct design the action space comprises search, lookup, and finish to help the agent interact with an external Wikipedia tool to retrieve necessary information. \textbf{ReAct-IM} uses Inner Monologue (IM)~\citep{huang2022inner} style prompting. Chain-of-thought prompting (\textbf{CoT})~\citep{wei2022chain} enhances the reasoning capabilities of LLMs by generating a sequence of intermediate reasoning steps, which can be viewed as a reasoning-only baseline of ReAct removing actions and observations. \textbf{CoT-SC} ~\citep{wang2022self-consistency,wang2022rationale} is a follow-up work of CoT, serving as a self-consistency baseline. All these methods use greedy decoding, except that BUTLER~\cite{micheli2021language} uses beam search.
% Most of the existing methods are prompting methods based on designing few-shot prompting examples to boost the pre-trained LLMs on the given task, which does not exactly match our LTC setting to train the model. We add an extra necessary baseline by fine-tuning the Llama-7B model with the collected instruction fine-tuning data~\ref{par:pretraining} for a fair comparison.
We compare the agents trained by \modelshort with existing prompting and instruction tuning methods, including ReAct~\citep{yao2023react}, ReAct-IM~\citep{huang2022inner}, CoT~\citep{wei2022chain}, CoT-SC~\citep{wang2022self-consistency,wang2022rationale}, BUTLER~\cite{micheli2021language}. The detailed of these baselines are described in Appendix~\ref{sec:appendix_d}. Most of these methods focus on few-shot prompting, and different pre-trained models are used. To ensure a fair comparison, we include the additional baselines named ReAct-Tuning and CoT-Tuning by fine-tuning the Llama-7B model using the collected trajectories as fine-tuning data. In addition, GPT-4 are not used in the test time, and all the results reported are obtained by the trained agent itself.
% In our implementation, we add an extra testing phase to evaluate the ability of the current agent on the test datasets. All the numbers reported in this paper are tested on the held-out test set and without using GPT-4 in the testing. 

% \input{table_main}

\subsection{Results}
% \subsection{decision-making}

\myparagraph{\textit{ALFWorld}}
% \begin{table}[t]
\begin{wraptable}{br}{0.58\textwidth}
% \begin{table}[ht]
\vspace{-15pt}
\centering
\small
\begin{minipage}{.98\linewidth}
    \centering

\resizebox{\columnwidth}{!}{%
\begin{tabular}{l  cccccc  c}
\toprule
Method  \textbackslash ~Task  & Pick & Clean & Heat & Cool & Look & Pick 2 & All \\ \midrule
% Act {\tiny(best of 6)} & 88 & 42 & 74 & 67 & 72 & \textbf{41} & 45 \\
ReAct {\tiny(avg)} & 65 & 39 & 83 & 76 & 55 & 24 & 57 \\ 
ReAct {\tiny(best of 6)}  & \textbf{92} & 58 & \textbf{96} & 86 & 78 & 41 & 71 \\
\midrule
ReAct-IM  {\tiny(avg)}       & 55 & 59 & 60 & 55 & 23 & 24 & 48 \\
ReAct-IM {\tiny(best of 6)}  & 62 & 68 & 87 & 57 & 39 & 33 & 53 \\ 
\midrule
BUTLER$_g$ {\tiny(best of 8)}  & 33 & 26 & 70 & 76 & 17 & 12 & 22 \\
BUTLER {\tiny(best of 8)}  & 46 & 39 & 74 & \textbf{100} & 22 & 24 & 37 \\
\midrule \midrule
ReAct-Tuning {\tiny(avg)}  & 83 & 91 & 91 & 90 & 72 & 8 & 77 \\
ReAct-Tuning {\tiny(best of 3)} & \textbf{92} & \textbf{97} & \textbf{96} & 95 & 78 & 24 & 78 \\
LTC  {\tiny(avg)}   & 89 & 91 & 93 & 97 & 96 & 67 & 90 \\
LTC  {\tiny(best of 3)}   & \textbf{92} & \textbf{97} & \textbf{96} & \textbf{100} & \textbf{100} & \textbf{76} & \textbf{91} \\
\bottomrule
\end{tabular}%
}
\vspace{-10pt}
\caption{
    AlfWorld success rates (\%) for 6 tasks. The results of the bottom block are obtained by fine-tuning Llama-7B model. 
    %BUTLER and BUTLER$_g$ results are from Table 4 of~\cite{shridhar2020alfworld}, and they are trained with DAgger~\cite{ross2011reduction}. ReAct and ReAct-IM results are from Table 3 of~\cite{yao2023react}.
    % All methods use greedy decoding, except that BUTLER~\cite{micheli2021language} uses beam search.
}
\label{tab:alfworld_detail}

\vspace{-10pt}
\end{minipage}%

% \hspace{5pt}
% \begin{minipage}{.23\linewidth}
%     \centering
% \resizebox{\columnwidth}{!}{%
% \begin{tabular}{c | cc}
% \toprule
% Method & Score & SR \\ \midrule
% Act & 62.3 & 30.1 \\
% ReAct & \textbf{66.6} & \textbf{40.0} \\ \midrule
% IL & 59.9 & 29.1 \\ 
% IL+RL & 62.4 & 28.7 \\ \midrule \midrule
% Human & \multirow{2}{*}{82.1}  & \multirow{2}{*}{59.6} \\
% Expert & & \\
% \bottomrule
% \end{tabular}%
% }
% \caption{
%     Score and success rate (SR) on Webshop. %IL/IL+RL taken from \cite{yao2022webshop}. 
% }
% \label{tab:alfworld}
% \end{minipage}%
% \vspace{-10pt}

% \end{table}
\end{wraptable}

As shown in \tbl{tab:alfworld_detail}, \modelshort outperforms the previous best methods\footnote{For \textit{ALFWorld}, ReAct and ReAct-IM results are from Table 3 of~\cite{yao2023react}. BUTLER and BUTLER$_g$ results are from Table 4 of~\cite{shridhar2020alfworld}, and they are trained with DAgger~\cite{ross2011reduction}.} on all of tasks of \textit{ALFWorld}. We can see that Instruction Fine-tuning is already a strong baseline outperforming others, yet our \modelshort achieves a success rate of 91\%, remarkably outperforming the best Instruction Tuning baseline (78\%).
Notably, on both Cool and Look tasks, \modelshort obtains a 100\% success rate.
% \begin{table}[t]
\begin{wraptable}{br}{0.42\textwidth}
% \begin{table}[ht]
\vspace{-10pt}

\begin{minipage}{.98\linewidth}

\centering
% \small

\resizebox{\columnwidth}{!}{%
\begin{tabular}{llc}
\toprule
    %  \multirow{2}{*}{\textbf{Model}} & \multirow{2}{*}{\textbf{Method\footnote{\tiny HotpotQA EM is 27.1, 28.9, 33.8 for Standard, CoT, CoT-SC in \cite{wang2022rationale}.}}} & \textbf{HotpotQA}   \\
    % & (EM) \\ 
    \textbf{Model} & \textbf{Method} & \textbf{EM score}   \\
    \midrule
    % \multirow{4}{*}{PaLM-540B} & Standard  & 28.7 \\
    \multirow{3}{*}{PaLM-540B} & CoT{\scriptsize~\citep{wei2022chain}} & 29.4 \\ 
    & CoT-SC{\scriptsize~\citep{wang2022self-consistency}} & 33.4\\
    % \midrule
    % &  Act  & 25.7 \\ 
    & ReAct{\scriptsize~\citep{yao2023react}}   & 27.4  \\
    & ReAct $\to$ CoT-SC & 35.1 \\
    \midrule
    GPT3-175B & ReAct   & 30.8 \\
    % & CoT-SC  $\to$ ReAct   & 34.2 \\
    % & ReAct $\to$ CoT-SC & \textbf{35.1}  \\ \midrule \midrule
    % & \textbf{Supervised SoTA\footnote{\tiny\citep{zhu2021adaptive,lewis2020retrieval}}} & 67.5 \\
    \midrule \midrule
    \multirow{2}{*}{PaLM-62B} & ReAct-Tuning   & 32.6 \\
    & CoT-Tuning   & 25.2 \\
    \midrule
    \multirow{2}{*}{PaLM-8B} & ReAct-Tuning   & 25.0 \\
    & CoT-Tuning   & 14.0 \\
    \midrule
    \multirow{2}{*}{Llama-7B} & ReAct-Tuning   & 28.2 \\
    & \modelshort {\tiny(single-agent monologue)}   & 31.0 \\
    & \modelshort {\tiny(multi-agent dialogue)}    & \textbf{33.2} \\
    \midrule
    \multirow{2}{*}{Llama2-13B} & ReAct-Tuning   & 33.8 \\
    & \modelshort {\tiny(multi-agent dialogue)}   & \textbf{35.8} \\

\bottomrule
\end{tabular}%
}
\caption{
EM scores on HotpotQA with prompt and tuning methods. Methods that use fine-tuning are marked by ``-Tuning''.
%The results of the bottom blocks are obtained by fine-tuning, while the others are prompting methods without fine-tuning from Table 1\&5 of~\cite{yao2023react}. PaLM-8b and PaLM-62B scores are estimates from Figure 3 of~\cite{yao2023react}.
% Prompting method results without fine-tuning are from Table 1\&5 of~\cite{yao2023react}. PaLM-8b and PaLM-62B scores are estimates from Figure 3 of~\cite{yao2023react}.
}
\label{tab:hotpotqa}
\vspace{-15pt}

% \end{table}

\end{minipage}

\end{wraptable}

Even on the hardest Pick Two \& Place task (e.g., “put two pencils in the drawer”), it achieves a decent 76\% success rate.
The Pick Two task requires the agent to perform two sequences of "pick and place" actions in one task, while keeping track of the desired type and the location.
The combined sequences and the need to remember the previous location make this task challenging.
This may be the reason why baselines achieve lower success rates on this task.
In contrast, our LTC agent, which further trains the agent with self-exploration significantly outperforms other agents. This underscores the effectiveness of the communication mechanism in LTC.
\myparagraph{\textit{HotpotQA}}

As shown in \tbl{tab:hotpotqa}, \modelshort outperforms the instruction tuning baseline\footnote{For HotPotQA, Prompting method results without fine-tuning are from Table 1\&5 of~\cite{yao2023react}. PaLM-8B and PaLM-62B scores are estimates from Figure 3 of~\cite{yao2023react}.} by 5\% on Exact Match (EM) score, and it even outperforms ReAct and CoT on their default settings. Note that ReAct and CoT use PaLM-540B and GPT3-175B as the pre-trained LM model, which is 77x and 25x larger than our the Llama-7B model we used. By sampling 21 CoT trajectories during inference and adopting the majority answer, CoT-SC is slightly better (0.2\%) than \modelshort, and their combined method ReAct $\to$ CoT-SC surpasses \modelshort by 1.9\%. Compared to other models with tuning, our Llama-7B based agent even obtains slightly better (0.6\%) performance than the ReAct-Tuning baseline with 9$\times$ larger PaLM-62B model.

% \begin{table}[t]
\begin{wraptable}{br}{0.45\textwidth}
% \begin{table}[ht]
\vspace{-20pt}
\centering
\small
\begin{minipage}{.98\linewidth}
    \centering

\resizebox{\columnwidth}{!}{%
\begin{tabular}{l  cccc}
\toprule
Method  \textbackslash ~\#players  & n=3 & n=4 & n=5 & overall  \\ \midrule
Llama-Tuning & 20.8 & 20.3 & 23.8 & 21.9  \\
Llama-LTC     & \textbf{22.9} & \textbf{23.4} & \textbf{27.5} & \textbf{25.0}  \\
\bottomrule
\end{tabular}%
}
\caption{
    Chameleon game winning rates (\%) of different numbers of players. 
    % At each game, one player is played by target evaluated model, and the others are played GPT-4.
}
\label{tab:chameleon}
\vspace{-20pt}

% \end{minipage}%

% \hspace{5pt}
% \begin{minipage}{.23\linewidth}
%     \centering
% \resizebox{\columnwidth}{!}{%
% \begin{tabular}{c | cc}
% \toprule
% Method & Score & SR \\ \midrule
% Act & 62.3 & 30.1 \\
% ReAct & \textbf{66.6} & \textbf{40.0} \\ \midrule
% IL & 59.9 & 29.1 \\ 
% IL+RL & 62.4 & 28.7 \\ \midrule \midrule
% Human & \multirow{2}{*}{82.1}  & \multirow{2}{*}{59.6} \\
% Expert & & \\
% \bottomrule
% \end{tabular}%
% }
% \caption{
%     Score and success rate (SR) on Webshop. %IL/IL+RL taken from \cite{yao2022webshop}. 
% }
% \label{tab:alfworld}
\end{minipage}%
% \vspace{-15pt}

% \end{table}
\end{wraptable}

\myparagraph{\textit{Chameleon}}

As shown in \tbl{tab:chameleon}, \modelshort outperforms the instruction tuning baselines by 3.1\% on winning rate against GPT-4 players. In the training, all the players are played by the same Llama2-7B model that we are training. While in the testing, to get the winning rate of our trained agent against GPT4, only 1 player is randomly picked to use our trained agent as backend, and other players are played by GPT4. 
We could see that the LTC agents winning rate improves with the increasing of number of players, we explain this by the more players, the higher chance that the GPT4 players carry the game.

\myparagraph{\textit{GSM8k}}
As shown in \tbl{tab:gsm8k}, \modelshort {\tiny(teacher-student dialogue)} outperforms the instruction fine-tuning baseline by 3.6\% on accuracy, and it surpasses the \modelshort {\tiny(single-agent monologue)} baseline, which does not use the reward and feedback from GPT-4.
% It also outperforms CoT and CoT-SC baselines with the same Llama-7B model.
% \begin{table}[t]
\begin{wraptable}{br}{0.42\textwidth}
% \begin{table}[ht]
\vspace{-10pt}
\centering
% \small

\begin{minipage}{.98\linewidth}

\centering
% \small

\resizebox{\columnwidth}{!}{%

\begin{tabular}{llc}
\toprule
    %  \multirow{2}{*}{\textbf{Model}} & \multirow{2}{*}{\textbf{Method\footnote{\tiny HotpotQA EM is 27.1, 28.9, 33.8 for Standard, CoT, CoT-SC in \cite{wang2022rationale}.}}} & \textbf{HotpotQA}   \\
    % & (EM) \\ 
    \textbf{Model} & \textbf{Method} & \textbf{Accuracy}   \\
    \midrule
    % \multirow{4}{*}{PaLM-540B} & Standard  & 28.7 \\
    \multirow{2}{*}{PaLM-540B} & CoT{\scriptsize~\citep{wei2022chain}} & 56.5 \\ 
    & CoT-SC{\scriptsize~\citep{wang2022self-consistency}} & 74.4\\
    % \midrule
    % &  Act  & 25.7 \\ 
 
    \midrule
    \multirow{2}{*}{GPT3-175B} & CoT{\scriptsize~\citep{wei2022chain}}   & 60.1 \\
    & CoT-SC{\scriptsize~\citep{wang2022self-consistency}} & 78.0\\
    % & CoT-SC  $\to$ ReAct   & 34.2 \\
    % & ReAct $\to$ CoT-SC & \textbf{35.1}  \\ \midrule \midrule
    % & \textbf{Supervised SoTA\footnote{\tiny\citep{zhu2021adaptive,lewis2020retrieval}}} & 67.5 \\
    \midrule 
    \multirow{2}{*}{Llama-7B} & CoT{\scriptsize~\citep{touvron2023llama}} & 11.0 \\
    & CoT-SC{\scriptsize~\citep{touvron2023llama}}    & 18.1 \\
    \midrule \midrule
    \multirow{3}{*}{Llama-7B} & CoT-Tuning   & 37.7 \\
    & \modelshort {\tiny(single-agent monologue)}   & 39.6 \\
    & \modelshort {\tiny(teacher-student dialogue)}   & \textbf{41.3} \\

\bottomrule
\end{tabular}%
}
\vspace{-10pt}
\caption{
Accuracy on GSM8k. The results of the bottom block are obtained by fine-tuning, while the others are prompting methods. 
% Accuracy on GSM8k. The results of the bottom block are obtained by fine-tuning LLaMA-7B model, while the others are prompting methods without fine-tuning. 
}
\label{tab:gsm8k}
\vspace{-25pt}

\end{minipage}

% \end{table}
\end{wraptable}

However, \modelshort underperforms CoT and CoT-SC with the much larger models (PaLM-540B and GPT3-175B).
This phenomenon is because numerical reasoning requires a larger model size and sufficient pretraining data, as observed in \cite{openai2023gpt4}.
Unfortunately, due to computational resource limitations, we can only train the relatively small Llama-7B model but were unable to train larger-scale models.
Nevertheless, we believe that exploring \modelshort with larger models is promising for future research.

\section{Discussion}
% \subsection{Model efficiency}
% \subsection{Data efficiency}

% \input{curves_ppo_interaction}
\myparagraph{Shortcuts}
One interesting observation is that the GPT-4 agent sometimes employs "shortcuts" to solve problems when serving as a teacher to generate new training data. These shortcuts rely on the internal knowledge acquired during its pretraining process. To illustrate this, we present a case study from \textit{HotpotQA} in  \fig{fig:shortcut_compare}. In this case, the GPT-4 agent quickly retrieves the answer by leveraging its memorized knowledge about the second entry after receiving the Wikipedia page of the first entry. On the other hand, the bottom of  \fig{fig:shortcut_compare} demonstrates a comparison with LLaMA-7B, which was trained using our LTC  method with the GPT-4 agent in the loop. LLaMA-7B does not employ shortcuts and instead performs a search for the second entry. This case study   demonstrates that communication mechanism in LTC provide additional benefits during learning, compared to soley relying on data generated by GPT-4.

\myparagraph{Efficiency} 
\begin{wraptable}{br}{0.48\textwidth}
% \begin{table}[ht]
% \vspace{-10pt}
\centering
\small

\begin{minipage}{.98\linewidth}

\centering
% \small

\resizebox{\columnwidth}{!}{%

% \begin{table}[ht]
% \centering
\begin{tabular}{lccc}
\toprule
\multirow{2}{*}{\textbf{Method}} & \textbf{GSM8k}  & \textbf{Hotpot-QA}  & \textbf{Alfworld}  \\ 
& {\tiny(CoT)}  & {\tiny(ReAct)}  & {\tiny(ReAct)} \\
% & \textbf{GSM8k}{\tiny(CoT)}  & \textbf{Hotpot-QA}{\tiny(ReAct)}  & \textbf{Alfworld}{\tiny(ReAct)} 
\hline
\textbf{ICL} & 836 & 1937 & 1744 \\ 
% \hdashline
\midrule
\textbf{LTC} & \textbf{107} & \textbf{167} & \textbf{189}  \\
\bottomrule
% \textbf{LLaMA + GPT4 Communication} & \textbf{41.0}\% & 30.1\% & -\%  \\
% \textbf{LLaMA + Self-feedback} & -\% & 30.1\% & -\%  \\
\end{tabular}
% \vspace{-10pt}
}
\caption{Average number of tokens of the input prompts on test sets. LTC does not use any few shot examples in the prompt, hence uses only a fraction of tokens compared to ICL. }
% \vspace{-10pt}
\label{tab:tokens}
% \end{table}

\end{minipage}

\end{wraptable} 
As mentioned above, prompting-based methods such as ReAct~\citep{yao2023react} and CoT~\citep{wei2022chain} use a subset of exemplary trajectories from the given task as few-shot prompts during inference. 
% In these works, the few-shot prompts serve as short-term memory in In-context learning (ICL), making the methods training-free.
However, these few-shot prompts are often long, which leads to increased inference cost and limited context length for user queries.
% , as the tokens from these examples might constitute a significant portion of the total input tokens for the target tasks. It is notably inefficient for specific tasks with a large number of queries directed at the same agent. 
As shown in \tbl{tab:tokens}, we compare the number of input tokens for each task. 
We compute the CoT prompts for \textbf{GSM8k}, and we use ReAct for the other two tasks. All the few-shot prompts are sourced from the original paper. As shown, our LTC agents used only 12.8\%, 8.6\%, and 10.8\% of the input tokens required by the ICL methods on the three tasks, respectively.  

% \begin{figure}[h]
%   \centering
%   \includegraphics[scale=0.4]{figures/curves_alfworld_ppo.png}
%   \caption{The acc curves of PPO training.}
%   \label{fig:curves_ppo}
% \end{figure}

% \begin{table}[t]
% \begin{wraptable}{br}{0.55\textwidth}
\begin{figure*}[t]
\vspace{-40pt}
\centering
\small
\begin{minipage}{.49\linewidth}
    \centering
    \includegraphics[scale=0.42]{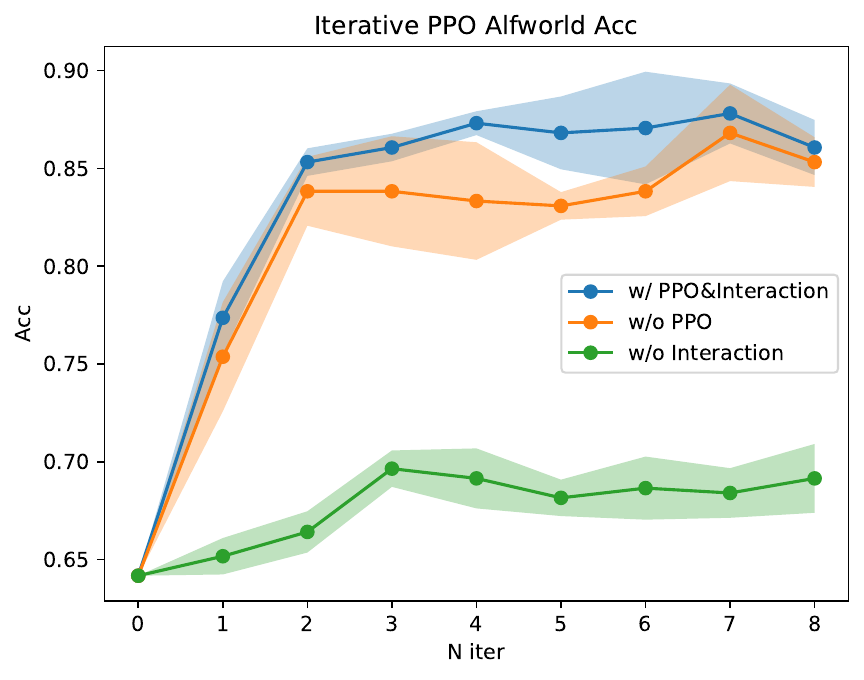}
    % \caption{The accuracy curves of PPO training.}
    \vspace{-10pt}
    \caption{The accuracy curves of training.}
    \label{fig:curves_ppo}
    
\end{minipage}%
% \hspace{10pt}
\begin{minipage}{.49\linewidth}
    \centering
    \includegraphics[scale=0.42]{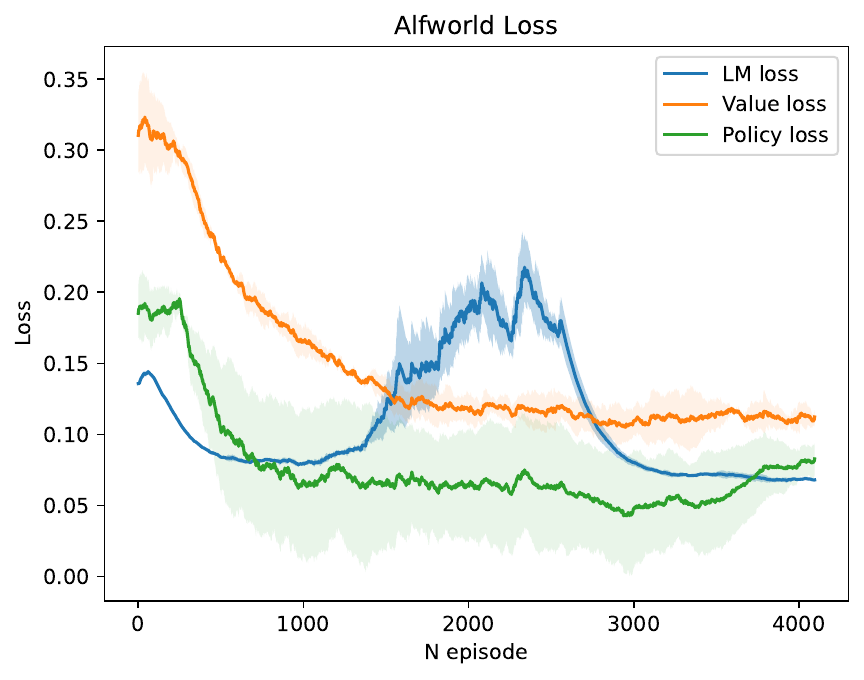}
    \vspace{-10pt}
    \caption{The loss curves of training.}
    % \caption{The loss curves of PPO training.}
    \label{fig:curves_ppo_loss}
    
\end{minipage}%
% \vspace{-10pt}
\end{figure*}

% \kuan{remove \fig{fig:shortcut} if space is limited}
% \input{fig_shortcut_compare}
\myparagraph{Ablation} 

We conducted ablation studies on the loss design of LTC.  \fig{fig:curves_ppo} illustrates the success rate of agents on the \textit{ALFWorld} dataset under different loss settings. Without using our communication pattern for interactions and merely sampling pre-collected instruction data for training, the improvement was limited. However, when we incorporated our communication pattern to gather data, the model's performance quickly surpassed 80\%. 
Furthermore, employing PPO loss to handle positive and negative samples separately resulted in faster and more significant improvement (blue line).
In  \fig{fig:curves_ppo_loss}, we present the separate curves of the three main losses during training. Initially, the LM loss showed a decreasing trend. Interestingly, as training iterations progressed, both the value loss and policy loss gradually decreased, which possibly causes the LM loss to increase temporarily. After the value loss and policy loss reached a certain threshold, the LM loss continued to decrease till convergence.

% \subsection{Shortcuts}
% \input{fig_shortcut}

% \input{curves_ppo_loss}
% \input{curves_ppo_dev}

\section{Conclusion}

% We present the LTC paradigm, a communication-based iterative learning framework to adapt LLM based agents to new tasks. We designed three structured communication patterns for common tasks such as decision-making, knowledge-intensive reasoning and numeric reasoning. These communication patterns serve as guideline for the interactions of LLM based agents with environments and other agents such as GPT-4 and human.

% . Our approach closes the iterative loop of self interact with environment or other agents, gather new exploration data, and continuously improve itself, with minimal human efforts.  Notably, we demonstrated LTC achieves consistent improvement across three different tasks: AlfWorld, HotpotQA and GSM8k in both performance and efficiency over ReAct and instruction tuning baseline.  With increasing capabilities of LLMs, our method takes a solid step towards efficiently adapting agents to new task with minimal human efforts. 

We introduced Learning-Through-Communication (LTC), a paradigm that adapts LLM agents to new tasks and environments via communication-based iterative learning.
Within this LTC framework, we have designed three communication modes for common tasks including decision-making, knowledge-intensive reasoning, and numeric reasoning.
These communication modes facilitate interactions between LLM agents and their environments, as well as other agents such as GPT-4 and humans.
The history of these interactions can be autonomously organized into training data for PPO training so that the agent can adapt to the new task.
Our approach represents a closed loop where the agent self-interacts with the environment or other agents, and learning to improve itself with minimal human intervention.
Empirically, we have demonstrated that LTC performs strongly in success rate and efficiency across four different tasks: AlfWorld, HotpotQA, Chameleon, and GSM8k.
It consistently outperforms existing LLM agent and instruction tuning baselines, showing the promise of the LTC paradigm in adapting LLM agents to new tasks and environments with minimal human effort. 
As for future work, we plan to explore more diverse communication patterns for different tasks, and involve the communication with human during the iterative learning process.
We will open source our code to facilitate further research in this line.
 % \zc{ Better to add one or two sentences about future work.} 

% By mimicking human-like learning interactions and seamlessly integrating with advanced reinforcement learning techniques like PPO, LTC promises to usher in a new era of model training that is both intuitive and effective. 

% \section{Impact Statement}
% This paper presents work whose goal is to advance the field of Machine Learning. There might be potential societal consequences of our work, none of which we feel must be specifically highlighted here.

% Acknowledgements should only appear in the accepted version.
% \section*{Acknowledgements}

% In the unusual situation where you want a paper to appear in the
% references without citing it in the main text, use \nocite
% \nocite{langley00}

\bibliography{reference}

\begin{thebibliography}{72}
\providecommand{\natexlab}[1]{#1}
\providecommand{\url}[1]{\texttt{#1}}
\expandafter\ifx\csname urlstyle\endcsname\relax
  \providecommand{\doi}[1]{doi: #1}\else
  \providecommand{\doi}{doi: \begingroup \urlstyle{rm}\Url}\fi

\bibitem[Akata et~al.(2023)Akata, Schulz, Coda-Forno, Oh, Bethge, and Schulz]{akata2023playing}
Elif Akata, Lion Schulz, Julian Coda-Forno, Seong~Joon Oh, Matthias Bethge, and Eric Schulz.
\newblock Playing repeated games with large language models.
\newblock \emph{arXiv preprint arXiv:2305.16867}, 2023.

\bibitem[AntonOsika(2023)]{AntonOsika2023}
AntonOsika.
\newblock gpt-engineer.
\newblock \url{https://github.com/AntonOsika/gpt-engineer}, 2023.
\newblock GitHub repository.

\bibitem[Bai et~al.(2022{\natexlab{a}})Bai, Jones, Ndousse, Askell, Chen, DasSarma, Drain, Fort, Ganguli, Henighan, et~al.]{bai2022training}
Yuntao Bai, Andy Jones, Kamal Ndousse, Amanda Askell, Anna Chen, Nova DasSarma, Dawn Drain, Stanislav Fort, Deep Ganguli, Tom Henighan, et~al.
\newblock Training a helpful and harmless assistant with reinforcement learning from human feedback.
\newblock \emph{arXiv preprint arXiv:2204.05862}, 2022{\natexlab{a}}.

\bibitem[Bai et~al.(2022{\natexlab{b}})Bai, Kadavath, Kundu, Askell, Kernion, Jones, Chen, Goldie, Mirhoseini, McKinnon, et~al.]{bai2022constitutional}
Yuntao Bai, Saurav Kadavath, Sandipan Kundu, Amanda Askell, Jackson Kernion, Andy Jones, Anna Chen, Anna Goldie, Azalia Mirhoseini, Cameron McKinnon, et~al.
\newblock Constitutional ai: Harmlessness from ai feedback.
\newblock \emph{arXiv preprint arXiv:2212.08073}, 2022{\natexlab{b}}.

\bibitem[Brown et~al.(2020)Brown, Mann, Ryder, Subbiah, Kaplan, Dhariwal, Neelakantan, Shyam, Sastry, Askell, Agarwal, Herbert-Voss, Krueger, Henighan, Child, Ramesh, Ziegler, Wu, Winter, Hesse, Chen, Sigler, Litwin, Gray, Chess, Clark, Berner, McCandlish, Radford, Sutskever, and Amodei]{brown2020language}
Tom Brown, Benjamin Mann, Nick Ryder, Melanie Subbiah, Jared~D Kaplan, Prafulla Dhariwal, Arvind Neelakantan, Pranav Shyam, Girish Sastry, Amanda Askell, Sandhini Agarwal, Ariel Herbert-Voss, Gretchen Krueger, Tom Henighan, Rewon Child, Aditya Ramesh, Daniel Ziegler, Jeffrey Wu, Clemens Winter, Chris Hesse, Mark Chen, Eric Sigler, Mateusz Litwin, Scott Gray, Benjamin Chess, Jack Clark, Christopher Berner, Sam McCandlish, Alec Radford, Ilya Sutskever, and Dario Amodei.
\newblock Language models are few-shot learners.
\newblock In H.~Larochelle, M.~Ranzato, R.~Hadsell, M.~F. Balcan, and H.~Lin (eds.), \emph{Advances in Neural Information Processing Systems}, volume~33, pp.\  1877--1901. Curran Associates, Inc., 2020.

\bibitem[Bubeck et~al.(2023)Bubeck, Chandrasekaran, Eldan, Gehrke, Horvitz, Kamar, Lee, Lee, Li, Lundberg, et~al.]{bubeck2023sparks}
S{\'e}bastien Bubeck, Varun Chandrasekaran, Ronen Eldan, Johannes Gehrke, Eric Horvitz, Ece Kamar, Peter Lee, Yin~Tat Lee, Yuanzhi Li, Scott Lundberg, et~al.
\newblock Sparks of artificial general intelligence: Early experiments with gpt-4.
\newblock \emph{arXiv preprint arXiv:2303.12712}, 2023.

\bibitem[Chiang et~al.(2023)Chiang, Li, Lin, Sheng, Wu, Zhang, Zheng, Zhuang, Zhuang, Gonzalez, et~al.]{chiang2023vicuna}
Wei-Lin Chiang, Zhuohan Li, Zi~Lin, Ying Sheng, Zhanghao Wu, Hao Zhang, Lianmin Zheng, Siyuan Zhuang, Yonghao Zhuang, Joseph~E Gonzalez, et~al.
\newblock Vicuna: An open-source chatbot impressing gpt-4 with 90\%* chatgpt quality.
\newblock \emph{See https://vicuna. lmsys. org (accessed 14 April 2023)}, 2023.

\bibitem[Chowdhery et~al.(2022)Chowdhery, Narang, Devlin, Bosma, Mishra, Roberts, Barham, Chung, Sutton, Gehrmann, et~al.]{chowdhery2022palm}
Aakanksha Chowdhery, Sharan Narang, Jacob Devlin, Maarten Bosma, Gaurav Mishra, Adam Roberts, Paul Barham, Hyung~Won Chung, Charles Sutton, Sebastian Gehrmann, et~al.
\newblock Palm: Scaling language modeling with pathways.
\newblock \emph{arXiv preprint arXiv:2204.02311}, 2022.

\bibitem[Chung et~al.(2022)Chung, Hou, Longpre, Zoph, Tay, Fedus, Li, Wang, Dehghani, Brahma, et~al.]{chung2022scaling}
Hyung~Won Chung, Le~Hou, Shayne Longpre, Barret Zoph, Yi~Tay, William Fedus, Eric Li, Xuezhi Wang, Mostafa Dehghani, Siddhartha Brahma, et~al.
\newblock Scaling instruction-finetuned language models.
\newblock \emph{arXiv preprint arXiv:2210.11416}, 2022.

\bibitem[Cobbe et~al.(2021)Cobbe, Kosaraju, Bavarian, Chen, Jun, Kaiser, Plappert, Tworek, Hilton, Nakano, Hesse, and Schulman]{cobbe2021gsm8k}
Karl Cobbe, Vineet Kosaraju, Mohammad Bavarian, Mark Chen, Heewoo Jun, Lukasz Kaiser, Matthias Plappert, Jerry Tworek, Jacob Hilton, Reiichiro Nakano, Christopher Hesse, and John Schulman.
\newblock Training verifiers to solve math word problems.
\newblock \emph{arXiv preprint arXiv:2110.14168}, 2021.

\bibitem[Deng et~al.(2023)Deng, Gu, Zheng, Chen, Stevens, Wang, Sun, and Su]{deng2023mind2web}
Xiang Deng, Yu~Gu, Boyuan Zheng, Shijie Chen, Samuel Stevens, Boshi Wang, Huan Sun, and Yu~Su.
\newblock Mind2web: Towards a generalist agent for the web, 2023.

\bibitem[Gao et~al.(2022)Gao, Madaan, Zhou, Alon, Liu, Yang, Callan, and Neubig]{gao2022pal}
Luyu Gao, Aman Madaan, Shuyan Zhou, Uri Alon, Pengfei Liu, Yiming Yang, Jamie Callan, and Graham Neubig.
\newblock Pal: Program-aided language models.
\newblock \emph{arXiv preprint arXiv:2211.10435}, 2022.

\bibitem[Hong et~al.(2023)Hong, Zheng, Chen, Cheng, Wang, Zhang, Wang, Yau, Lin, Zhou, Ran, Xiao, and Wu]{hong2023metagpt}
Sirui Hong, Xiawu Zheng, Jonathan Chen, Yuheng Cheng, Jinlin Wang, Ceyao Zhang, Zili Wang, Steven Ka~Shing Yau, Zijuan Lin, Liyang Zhou, Chenyu Ran, Lingfeng Xiao, and Chenglin Wu.
\newblock Metagpt: Meta programming for multi-agent collaborative framework, 2023.

\bibitem[Honovich et~al.(2022)Honovich, Scialom, Levy, and Schick]{honovich2022unnatural}
Or~Honovich, Thomas Scialom, Omer Levy, and Timo Schick.
\newblock Unnatural instructions: Tuning language models with (almost) no human labor.
\newblock \emph{arXiv preprint arXiv:2212.09689}, 2022.

\bibitem[Hu et~al.(2021)Hu, Shen, Wallis, Allen-Zhu, Li, Wang, Wang, and Chen]{hu2021lora}
Edward~J Hu, Yelong Shen, Phillip Wallis, Zeyuan Allen-Zhu, Yuanzhi Li, Shean Wang, Lu~Wang, and Weizhu Chen.
\newblock Lora: Low-rank adaptation of large language models.
\newblock \emph{arXiv preprint arXiv:2106.09685}, 2021.

\bibitem[Huang et~al.(2022{\natexlab{a}})Huang, Gu, Hou, Wu, Wang, Yu, and Han]{huang2022large}
Jiaxin Huang, Shixiang~Shane Gu, Le~Hou, Yuexin Wu, Xuezhi Wang, Hongkun Yu, and Jiawei Han.
\newblock Large language models can self-improve, 2022{\natexlab{a}}.

\bibitem[Huang et~al.(2022{\natexlab{b}})Huang, Xia, Xiao, Chan, Liang, Florence, Zeng, Tompson, Mordatch, Chebotar, Sermanet, Brown, Jackson, Luu, Levine, Hausman, and Ichter]{huang2022inner}
Wenlong Huang, Fei Xia, Ted Xiao, Harris Chan, Jacky Liang, Pete Florence, Andy Zeng, Jonathan Tompson, Igor Mordatch, Yevgen Chebotar, Pierre Sermanet, Noah Brown, Tomas Jackson, Linda Luu, Sergey Levine, Karol Hausman, and Brian Ichter.
\newblock Inner monologue: Embodied reasoning through planning with language models.
\newblock In \emph{arXiv preprint arXiv:2207.05608}, 2022{\natexlab{b}}.

\bibitem[Izacard et~al.(2022)Izacard, Lewis, Lomeli, Hosseini, Petroni, Schick, Dwivedi-Yu, Joulin, Riedel, and Grave]{izacard2022few}
Gautier Izacard, Patrick Lewis, Maria Lomeli, Lucas Hosseini, Fabio Petroni, Timo Schick, Jane Dwivedi-Yu, Armand Joulin, Sebastian Riedel, and Edouard Grave.
\newblock Few-shot learning with retrieval augmented language models.
\newblock \emph{arXiv preprint arXiv:2208.03299}, 2022.

\bibitem[Lee et~al.(2023)Lee, Phatale, Mansoor, Lu, Mesnard, Bishop, Carbune, and Rastogi]{lee2023rlaif}
Harrison Lee, Samrat Phatale, Hassan Mansoor, Kellie Lu, Thomas Mesnard, Colton Bishop, Victor Carbune, and Abhinav Rastogi.
\newblock Rlaif: Scaling reinforcement learning from human feedback with ai feedback.
\newblock \emph{arXiv preprint arXiv:2309.00267}, 2023.

\bibitem[Leike et~al.(2018)Leike, Krueger, Everitt, Martic, Maini, and Legg]{leike2018scalable}
Jan Leike, David Krueger, Tom Everitt, Miljan Martic, Vishal Maini, and Shane Legg.
\newblock Scalable agent alignment via reward modeling: a research direction, 2018.

\bibitem[Li et~al.(2023)Li, Hammoud, Itani, Khizbullin, and Ghanem]{li2023camel}
Guohao Li, Hasan Abed Al~Kader Hammoud, Hani Itani, Dmitrii Khizbullin, and Bernard Ghanem.
\newblock Camel: Communicative agents for "mind" exploration of large scale language model society, 2023.

\bibitem[Lin et~al.(2023)Lin, Fu, Yang, Ammanabrolu, Brahman, Huang, Bhagavatula, Choi, and Ren]{Lin2023SwiftSageAG}
Bill~Yuchen Lin, Yicheng Fu, Karina Yang, Prithviraj Ammanabrolu, Faeze Brahman, Shiyu Huang, Chandra Bhagavatula, Yejin Choi, and Xiang Ren.
\newblock Swiftsage: A generative agent with fast and slow thinking for complex interactive tasks.
\newblock \emph{ArXiv preprint}, abs/2305.17390, 2023.
\newblock URL \url{https://arxiv.org/abs/2305.17390}.

\bibitem[Liu et~al.(2023{\natexlab{a}})Liu, Yang, Jia, Zhang, Zhou, Dai, Yang, and Vosoughi]{liu2023training}
Ruibo Liu, Ruixin Yang, Chenyan Jia, Ge~Zhang, Denny Zhou, Andrew~M Dai, Diyi Yang, and Soroush Vosoughi.
\newblock Training socially aligned language models in simulated human society.
\newblock \emph{arXiv preprint arXiv:2305.16960}, 2023{\natexlab{a}}.

\bibitem[Liu et~al.(2023{\natexlab{b}})Liu, Yu, Zhang, Xu, Lei, Lai, Gu, Ding, Men, Yang, Zhang, Deng, Zeng, Du, Zhang, Shen, Zhang, Su, Sun, Huang, Dong, and Tang]{liu2023agentbench}
Xiao Liu, Hao Yu, Hanchen Zhang, Yifan Xu, Xuanyu Lei, Hanyu Lai, Yu~Gu, Hangliang Ding, Kaiwen Men, Kejuan Yang, Shudan Zhang, Xiang Deng, Aohan Zeng, Zhengxiao Du, Chenhui Zhang, Sheng Shen, Tianjun Zhang, Yu~Su, Huan Sun, Minlie Huang, Yuxiao Dong, and Jie Tang.
\newblock Agentbench: Evaluating llms as agents.
\newblock \emph{arXiv preprint arXiv: 2308.03688}, 2023{\natexlab{b}}.

\bibitem[Longpre et~al.(2023)Longpre, Hou, Vu, Webson, Chung, Tay, Zhou, Le, Zoph, Wei, et~al.]{longpre2023flan}
Shayne Longpre, Le~Hou, Tu~Vu, Albert Webson, Hyung~Won Chung, Yi~Tay, Denny Zhou, Quoc~V Le, Barret Zoph, Jason Wei, et~al.
\newblock The flan collection: Designing data and methods for effective instruction tuning.
\newblock \emph{arXiv preprint arXiv:2301.13688}, 2023.

\bibitem[Loshchilov \& Hutter(2017)Loshchilov and Hutter]{loshchilov2017decoupled}
Ilya Loshchilov and Frank Hutter.
\newblock Decoupled weight decay regularization.
\newblock \emph{arXiv preprint arXiv:1711.05101}, 2017.

\bibitem[Lu et~al.(2023)Lu, Peng, Cheng, Galley, Chang, Wu, Zhu, and Gao]{lu2023chameleon}
Pan Lu, Baolin Peng, Hao Cheng, Michel Galley, Kai-Wei Chang, Ying~Nian Wu, Song-Chun Zhu, and Jianfeng Gao.
\newblock Chameleon: Plug-and-play compositional reasoning with large language models.
\newblock \emph{arXiv preprint arXiv:2304.09842}, 2023.

\bibitem[Micheli \& Fleuret(2021)Micheli and Fleuret]{micheli2021language}
Vincent Micheli and Fran{\c{c}}ois Fleuret.
\newblock Language models are few-shot butlers.
\newblock \emph{arXiv preprint arXiv:2104.07972}, 2021.

\bibitem[Mnih et~al.(2016)Mnih, Badia, Mirza, Graves, Lillicrap, Harley, Silver, and Kavukcuoglu]{mnih2016asynchronous}
Volodymyr Mnih, Adria~Puigdomenech Badia, Mehdi Mirza, Alex Graves, Timothy Lillicrap, Tim Harley, David Silver, and Koray Kavukcuoglu.
\newblock Asynchronous methods for deep reinforcement learning.
\newblock In \emph{International conference on machine learning}, pp.\  1928--1937. PMLR, 2016.

\bibitem[Muennighoff et~al.(2022)Muennighoff, Wang, Sutawika, Roberts, Biderman, Scao, Bari, Shen, Yong, Schoelkopf, et~al.]{muennighoff2022crosslingual}
Niklas Muennighoff, Thomas Wang, Lintang Sutawika, Adam Roberts, Stella Biderman, Teven~Le Scao, M~Saiful Bari, Sheng Shen, Zheng-Xin Yong, Hailey Schoelkopf, et~al.
\newblock Crosslingual generalization through multitask finetuning.
\newblock \emph{arXiv preprint arXiv:2211.01786}, 2022.

\bibitem[OpenAI(2022)]{chatgpt}
OpenAI.
\newblock Chatgpt.
\newblock Online, 2022.
\newblock URL \url{https://openai.com/blog/chatgpt}.

\bibitem[OpenAI(2023)]{openai2023gpt4}
OpenAI.
\newblock Gpt-4 technical report.
\newblock \emph{OpenAI blog}, 2023.

\bibitem[Ouyang et~al.(2022)Ouyang, Wu, Jiang, Almeida, Wainwright, Mishkin, Zhang, Agarwal, Slama, Ray, et~al.]{ouyang2022training}
Long Ouyang, Jeffrey Wu, Xu~Jiang, Diogo Almeida, Carroll Wainwright, Pamela Mishkin, Chong Zhang, Sandhini Agarwal, Katarina Slama, Alex Ray, et~al.
\newblock Training language models to follow instructions with human feedback.
\newblock \emph{Advances in Neural Information Processing Systems}, 35:\penalty0 27730--27744, 2022.

\bibitem[Park et~al.(2023)Park, O'Brien, Cai, Morris, Liang, and Bernstein]{Park2023GenerativeAgents}
Joon~Sung Park, Joseph~C. O'Brien, Carrie~J. Cai, Meredith~Ringel Morris, Percy Liang, and Michael~S. Bernstein.
\newblock Generative agents: Interactive simulacra of human behavior.
\newblock In \emph{In the 36th Annual ACM Symposium on User Interface Software and Technology (UIST '23)}, UIST '23, New York, NY, USA, 2023. Association for Computing Machinery.

\bibitem[Peng et~al.(2023)Peng, Li, He, Galley, and Gao]{peng2023instruction}
Baolin Peng, Chunyuan Li, Pengcheng He, Michel Galley, and Jianfeng Gao.
\newblock Instruction tuning with gpt-4.
\newblock \emph{arXiv preprint arXiv:2304.03277}, 2023.

\bibitem[Qian et~al.(2023)Qian, Cong, Yang, Chen, Su, Xu, Liu, and Sun]{qian2023communicative}
Chen Qian, Xin Cong, Cheng Yang, Weize Chen, Yusheng Su, Juyuan Xu, Zhiyuan Liu, and Maosong Sun.
\newblock Communicative agents for software development.
\newblock \emph{arXiv preprint arXiv:2307.07924}, 2023.

\bibitem[Qin et~al.(2023)Qin, Zhang, Zhang, Chen, Yasunaga, and Yang]{qin2023chatgpt}
Chengwei Qin, Aston Zhang, Zhuosheng Zhang, Jiaao Chen, Michihiro Yasunaga, and Diyi Yang.
\newblock Is chatgpt a general-purpose natural language processing task solver?, 2023.

\bibitem[Radford et~al.(2019)Radford, Wu, Child, Luan, Amodei, and Sutskever]{radford2019language}
Alec Radford, Jeff Wu, Rewon Child, David Luan, Dario Amodei, and Ilya Sutskever.
\newblock Language models are unsupervised multitask learners.
\newblock \emph{OpenAI blog}, 1\penalty0 (8):\penalty0 9, 2019.

\bibitem[Rafailov et~al.(2023)Rafailov, Sharma, Mitchell, Ermon, Manning, and Finn]{rafailov2023direct}
Rafael Rafailov, Archit Sharma, Eric Mitchell, Stefano Ermon, Christopher~D. Manning, and Chelsea Finn.
\newblock Direct preference optimization: Your language model is secretly a reward model, 2023.

\bibitem[Raffel et~al.(2020)Raffel, Shazeer, Roberts, Lee, Narang, Matena, Zhou, Li, and Liu]{2020t5}
Colin Raffel, Noam Shazeer, Adam Roberts, Katherine Lee, Sharan Narang, Michael Matena, Yanqi Zhou, Wei Li, and Peter~J. Liu.
\newblock Exploring the limits of transfer learning with a unified text-to-text transformer.
\newblock \emph{Journal of Machine Learning Research}, 21\penalty0 (140):\penalty0 1--67, 2020.

\bibitem[Ross et~al.(2011)Ross, Gordon, and Bagnell]{ross2011reduction}
St{\'e}phane Ross, Geoffrey Gordon, and Drew Bagnell.
\newblock A reduction of imitation learning and structured prediction to no-regret online learning.
\newblock In \emph{Proceedings of the fourteenth international conference on artificial intelligence and statistics}, pp.\  627--635. JMLR Workshop and Conference Proceedings, 2011.

\bibitem[Santacroce et~al.(2023)Santacroce, Lu, Yu, Li, and Shen]{santacroce2023efficient}
Michael Santacroce, Yadong Lu, Han Yu, Yuanzhi Li, and Yelong Shen.
\newblock Efficient rlhf: Reducing the memory usage of ppo, 2023.

\bibitem[Scao et~al.(2022)Scao, Fan, Akiki, Pavlick, Ili{\'c}, Hesslow, Castagn{\'e}, Luccioni, Yvon, Gall{\'e}, et~al.]{scao2022bloom}
Teven~Le Scao, Angela Fan, Christopher Akiki, Ellie Pavlick, Suzana Ili{\'c}, Daniel Hesslow, Roman Castagn{\'e}, Alexandra~Sasha Luccioni, Fran{\c{c}}ois Yvon, Matthias Gall{\'e}, et~al.
\newblock Bloom: A 176b-parameter open-access multilingual language model.
\newblock \emph{arXiv preprint arXiv:2211.05100}, 2022.

\bibitem[Schick et~al.(2023)Schick, Dwivedi-Yu, Dess{\`\i}, Raileanu, Lomeli, Zettlemoyer, Cancedda, and Scialom]{schick2023toolformer}
Timo Schick, Jane Dwivedi-Yu, Roberto Dess{\`\i}, Roberta Raileanu, Maria Lomeli, Luke Zettlemoyer, Nicola Cancedda, and Thomas Scialom.
\newblock Toolformer: Language models can teach themselves to use tools.
\newblock \emph{arXiv preprint arXiv:2302.04761}, 2023.

\bibitem[Schulman et~al.(2017)Schulman, Wolski, Dhariwal, Radford, and Klimov]{schulman2017proximal}
John Schulman, Filip Wolski, Prafulla Dhariwal, Alec Radford, and Oleg Klimov.
\newblock Proximal policy optimization algorithms, 2017.

\bibitem[Shi et~al.(2023)Shi, Min, Yasunaga, Seo, James, Lewis, Zettlemoyer, and Yih]{shi2023replug}
Weijia Shi, Sewon Min, Michihiro Yasunaga, Minjoon Seo, Rich James, Mike Lewis, Luke Zettlemoyer, and Wen-tau Yih.
\newblock Replug: Retrieval-augmented black-box language models.
\newblock \emph{arXiv preprint arXiv:2301.12652}, 2023.

\bibitem[Shinn et~al.(2023)Shinn, Cassano, Labash, Gopinath, Narasimhan, and Yao]{shinn2023reflexion}
Noah Shinn, Federico Cassano, Beck Labash, Ashwin Gopinath, Karthik Narasimhan, and Shunyu Yao.
\newblock Reflexion: Language agents with verbal reinforcement learning, 2023.

\bibitem[Shridhar et~al.(2020{\natexlab{a}})Shridhar, Thomason, Gordon, Bisk, Han, Mottaghi, Zettlemoyer, and Fox]{shridhar2020alfred}
Mohit Shridhar, Jesse Thomason, Daniel Gordon, Yonatan Bisk, Winson Han, Roozbeh Mottaghi, Luke Zettlemoyer, and Dieter Fox.
\newblock Alfred: A benchmark for interpreting grounded instructions for everyday tasks.
\newblock In \emph{Proceedings of the IEEE/CVF conference on computer vision and pattern recognition}, pp.\  10740--10749, 2020{\natexlab{a}}.

\bibitem[Shridhar et~al.(2020{\natexlab{b}})Shridhar, Yuan, C{\^o}t{\'e}, Bisk, Trischler, and Hausknecht]{shridhar2020alfworld}
Mohit Shridhar, Xingdi Yuan, Marc-Alexandre C{\^o}t{\'e}, Yonatan Bisk, Adam Trischler, and Matthew Hausknecht.
\newblock Alfworld: Aligning text and embodied environments for interactive learning.
\newblock \emph{arXiv preprint arXiv:2010.03768}, 2020{\natexlab{b}}.

\bibitem[Significant-Gravitas(2023)]{Gravitas2023}
Significant-Gravitas.
\newblock Autogpt.
\newblock \url{https://github.com/Significant-Gravitas/AutoGPT}, 2023.
\newblock GitHub repository.

\bibitem[Stiennon et~al.(2020)Stiennon, Ouyang, Wu, Ziegler, Lowe, Voss, Radford, Amodei, and Christiano]{stiennon2020learning}
Nisan Stiennon, Long Ouyang, Jeffrey Wu, Daniel Ziegler, Ryan Lowe, Chelsea Voss, Alec Radford, Dario Amodei, and Paul~F Christiano.
\newblock Learning to summarize with human feedback.
\newblock \emph{Advances in Neural Information Processing Systems}, 33:\penalty0 3008--3021, 2020.

\bibitem[Sumers et~al.(2023)Sumers, Yao, Narasimhan, and Griffiths]{sumers2023cognitive}
Theodore Sumers, Shunyu Yao, Karthik Narasimhan, and Thomas~L Griffiths.
\newblock Cognitive architectures for language agents.
\newblock \emph{arXiv preprint arXiv:2309.02427}, 2023.

\bibitem[Talebirad \& Nadiri(2023)Talebirad and Nadiri]{talebirad2023multi}
Yashar Talebirad and Amirhossein Nadiri.
\newblock Multi-agent collaboration: Harnessing the power of intelligent llm agents.
\newblock \emph{arXiv preprint arXiv:2306.03314}, 2023.

\bibitem[Taori et~al.(2023)Taori, Gulrajani, Zhang, Dubois, Li, Guestrin, Liang, and Hashimoto]{taori2023alpaca}
Rohan Taori, Ishaan Gulrajani, Tianyi Zhang, Yann Dubois, Xuechen Li, Carlos Guestrin, Percy Liang, and Tatsunori~B Hashimoto.
\newblock Alpaca: A strong, replicable instruction-following model.
\newblock \emph{Stanford Center for Research on Foundation Models. https://crfm. stanford. edu/2023/03/13/alpaca. html}, 3\penalty0 (6):\penalty0 7, 2023.

\bibitem[Touvron et~al.(2023)Touvron, Lavril, Izacard, Martinet, Lachaux, Lacroix, Rozi{\`e}re, Goyal, Hambro, Azhar, et~al.]{touvron2023llama}
Hugo Touvron, Thibaut Lavril, Gautier Izacard, Xavier Martinet, Marie-Anne Lachaux, Timoth{\'e}e Lacroix, Baptiste Rozi{\`e}re, Naman Goyal, Eric Hambro, Faisal Azhar, et~al.
\newblock Llama: Open and efficient foundation language models.
\newblock \emph{arXiv preprint arXiv:2302.13971}, 2023.

\bibitem[Wang et~al.(2022{\natexlab{a}})Wang, Jansen, C{\^o}t{\'e}, and Ammanabrolu]{scienceworld2022}
Ruoyao Wang, Peter Jansen, Marc-Alexandre C{\^o}t{\'e}, and Prithviraj Ammanabrolu.
\newblock Scienceworld: Is your agent smarter than a 5th grader?, 2022{\natexlab{a}}.
\newblock URL \url{https://arxiv.org/abs/2203.07540}.

\bibitem[Wang et~al.(2022{\natexlab{b}})Wang, Li, and Ji]{wang2022code4struct}
Xingyao Wang, Sha Li, and Heng Ji.
\newblock Code4struct: Code generation for few-shot structured prediction from natural language.
\newblock \emph{arXiv preprint arXiv:2210.12810}, 2022{\natexlab{b}}.

\bibitem[Wang et~al.(2022{\natexlab{c}})Wang, Wei, Schuurmans, Le, Chi, Narang, Chowdhery, and Zhou]{wang2022self-consistency}
Xuezhi Wang, Jason Wei, Dale Schuurmans, Quoc Le, Ed~Chi, Sharan Narang, Aakanksha Chowdhery, and Denny Zhou.
\newblock Self-consistency improves chain of thought reasoning in language models, 2022{\natexlab{c}}.
\newblock URL \url{https://arxiv.org/abs/2203.11171}.

\bibitem[Wang et~al.(2022{\natexlab{d}})Wang, Wei, Schuurmans, Le, Chi, and Zhou]{wang2022rationale}
Xuezhi Wang, Jason Wei, Dale Schuurmans, Quoc Le, Ed~Chi, and Denny Zhou.
\newblock Rationale-augmented ensembles in language models.
\newblock \emph{arXiv preprint arXiv:2207.00747}, 2022{\natexlab{d}}.

\bibitem[Wang et~al.(2022{\natexlab{e}})Wang, Kordi, Mishra, Liu, Smith, Khashabi, and Hajishirzi]{wang2022self}
Yizhong Wang, Yeganeh Kordi, Swaroop Mishra, Alisa Liu, Noah~A Smith, Daniel Khashabi, and Hannaneh Hajishirzi.
\newblock Self-instruct: Aligning language model with self generated instructions.
\newblock \emph{arXiv preprint arXiv:2212.10560}, 2022{\natexlab{e}}.

\bibitem[Wang et~al.(2022{\natexlab{f}})Wang, Mishra, Alipoormolabashi, Kordi, Mirzaei, Arunkumar, Ashok, Dhanasekaran, Naik, Stap, et~al.]{wang2022super}
Yizhong Wang, Swaroop Mishra, Pegah Alipoormolabashi, Yeganeh Kordi, Amirreza Mirzaei, Anjana Arunkumar, Arjun Ashok, Arut~Selvan Dhanasekaran, Atharva Naik, David Stap, et~al.
\newblock Super-naturalinstructions: Generalization via declarative instructions on 1600+ nlp tasks.
\newblock \emph{arXiv preprint arXiv:2204.07705}, 2022{\natexlab{f}}.

\bibitem[Wang et~al.(2023)Wang, Mao, Wu, Ge, Wei, and Ji]{wang2023unleashing}
Zhenhailong Wang, Shaoguang Mao, Wenshan Wu, Tao Ge, Furu Wei, and Heng Ji.
\newblock Unleashing cognitive synergy in large language models: A task-solving agent through multi-persona self-collaboration.
\newblock \emph{arXiv preprint arXiv:2307.05300}, 2023.

\bibitem[Wei et~al.(2022{\natexlab{a}})Wei, Tay, Bommasani, Raffel, Zoph, Borgeaud, Yogatama, Bosma, Zhou, Metzler, Chi, Hashimoto, Vinyals, Liang, Dean, and Fedus]{wei2022emergent}
Jason Wei, Yi~Tay, Rishi Bommasani, Colin Raffel, Barret Zoph, Sebastian Borgeaud, Dani Yogatama, Maarten Bosma, Denny Zhou, Donald Metzler, Ed~H. Chi, Tatsunori Hashimoto, Oriol Vinyals, Percy Liang, Jeff Dean, and William Fedus.
\newblock Emergent abilities of large language models, 2022{\natexlab{a}}.

\bibitem[Wei et~al.(2022{\natexlab{b}})Wei, Wang, Schuurmans, Bosma, Chi, Le, and Zhou]{wei2022chain}
Jason Wei, Xuezhi Wang, Dale Schuurmans, Maarten Bosma, Ed~Chi, Quoc Le, and Denny Zhou.
\newblock Chain of thought prompting elicits reasoning in large language models.
\newblock \emph{arXiv preprint arXiv:2201.11903}, 2022{\natexlab{b}}.

\bibitem[Weng(2023)]{weng2023prompt}
Lilian Weng.
\newblock Llm-powered autonomous agents.
\newblock \emph{lilianweng.github.io}, Jun 2023.
\newblock URL \url{https://lilianweng.github.io/posts/2023-06-23-agent/}.

\bibitem[Williams(1992)]{williams1992simple}
Ronald~J Williams.
\newblock Simple statistical gradient-following algorithms for connectionist reinforcement learning.
\newblock \emph{Machine learning}, 8:\penalty0 229--256, 1992.

\bibitem[Wu et~al.(2023{\natexlab{a}})Wu, Bansal, Zhang, Wu, Zhang, Zhu, Li, Jiang, Zhang, and Wang]{wu2023autogen}
Qingyun Wu, Gagan Bansal, Jieyu Zhang, Yiran Wu, Shaokun Zhang, Erkang Zhu, Beibin Li, Li~Jiang, Xiaoyun Zhang, and Chi Wang.
\newblock Autogen: Enabling next-gen llm applications via multi-agent conversation framework.
\newblock \emph{arXiv preprint arXiv:2308.08155}, 2023{\natexlab{a}}.

\bibitem[Wu et~al.(2023{\natexlab{b}})Wu, Jiang, Khan, Fu, Ruis, Grefenstette, and Rocktäschel]{ChatArena}
Yuxiang Wu, Zhengyao Jiang, Akbir Khan, Yao Fu, Laura Ruis, Edward Grefenstette, and Tim Rocktäschel.
\newblock Chatarena: Multi-agent language game environments for large language models.
\newblock \url{https://github.com/chatarena/chatarena}, 2023{\natexlab{b}}.

\bibitem[Xu et~al.(2023)Xu, Peng, Lei, Mukherjee, Liu, and Xu]{xu2023rewoo}
Binfeng Xu, Zhiyuan Peng, Bowen Lei, Subhabrata Mukherjee, Yuchen Liu, and Dongkuan Xu.
\newblock Rewoo: Decoupling reasoning from observations for efficient augmented language models.
\newblock \emph{arXiv preprint arXiv:2305.18323}, 2023.

\bibitem[Yang et~al.(2018)Yang, Qi, Zhang, Bengio, Cohen, Salakhutdinov, and Manning]{yang2018hotpotqa}
Zhilin Yang, Peng Qi, Saizheng Zhang, Yoshua Bengio, William~W. Cohen, Ruslan Salakhutdinov, and Christopher~D. Manning.
\newblock {HotpotQA}: A dataset for diverse, explainable multi-hop question answering.
\newblock In \emph{Conference on Empirical Methods in Natural Language Processing ({EMNLP})}, 2018.

\bibitem[Yao et~al.(2023)Yao, Zhao, Yu, Du, Shafran, Narasimhan, and Cao]{yao2023react}
Shunyu Yao, Jeffrey Zhao, Dian Yu, Nan Du, Izhak Shafran, Karthik~R Narasimhan, and Yuan Cao.
\newblock React: Synergizing reasoning and acting in language models.
\newblock In \emph{The Eleventh International Conference on Learning Representations}, 2023.
\newblock URL \url{https://openreview.net/forum?id=WE_vluYUL-X}.

\bibitem[yoheinakajima(2023)]{yoheinakajima2023}
yoheinakajima.
\newblock Babyagi.
\newblock \url{https://github.com/yoheinakajima/babyagi}, 2023.
\newblock GitHub repository.

\end{thebibliography}
\bibliographystyle{colm2024_conference}

%%%%%%%%%%%%%%%%%%%%%%%%%%%%%%%%%%%%%%%%%%%%%%%%%%%%%%%%%%%%%%%%%%%%%%%%%%%%%%%
%%%%%%%%%%%%%%%%%%%%%%%%%%%%%%%%%%%%%%%%%%%%%%%%%%%%%%%%%%%%%%%%%%%%%%%%%%%%%%%
% APPENDIX
%%%%%%%%%%%%%%%%%%%%%%%%%%%%%%%%%%%%%%%%%%%%%%%%%%%%%%%%%%%%%%%%%%%%%%%%%%%%%%%
%%%%%%%%%%%%%%%%%%%%%%%%%%%%%%%%%%%%%%%%%%%%%%%%%%%%%%%%%%%%%%%%%%%%%%%%%%%%%%%
\newpage
\appendix
\onecolumn

\section{Appendix}
% You may include other additional sections here.
% \input{curves_lm_loss}
% \input{curves_vs_if}
% \input{fig_shortcut_compare}
% \begin{wrapfigure}{R}{.5\textwidth}
% \begin{minipage}{.5\textwidth}
% % \vspace{-8mm}
\begin{algorithm}[t]

% \algcomment{\fontsize{7.2pt}{0em}\selectfont \texttt{bmm}: batch matrix multiplication; \texttt{mm}: matrix multiplication; \texttt{cat}: concatenation.
%\vspace{-1.em}
% }
\definecolor{codeblue}{rgb}{0.25,0.5,0.5}
\lstset{
  backgroundcolor=\color{white},
  basicstyle=\fontsize{7.2pt}{7.2pt}\ttfamily\selectfont,
  columns=fullflexible,
  breaklines=true,
  captionpos=b,
  commentstyle=\fontsize{7.2pt}{7.2pt}\color{codeblue},
  keywordstyle=\fontsize{7.2pt}{7.2pt},
%  frame=tb,
}
\begin{lstlisting}[language=python]
# agent: LLaMA agent
# input:  Task description 
# output: S  = (T, M, R)

# initialization
T, M, R = [input], [0], [0]

i = 0
while i < max_steps:
    T += ["think:"]
    thought = agent.api(T)
    T.append(thought)
    M.append(1) # agent message mask
    R.append(0)
    

    T += ["act:"]
    action = agent.api(T)
    T.append(action)
    M.append(1) # agent message mask
    R.append(0)

    response = env.excute(action)
    reward = parse(response)
    T.append(response)
    M.append(0) # system message mask
    R.append(reward)
    
    i += 1
    if reward != 0:
        break
S  = (T, M, R)   
return S
\end{lstlisting}
\caption{The Python-style algorithm to demonstrate Monologue pattern}
\label{alg:monologue}
\end{algorithm}
% \end{minipage}
% \vspace{-14pt}
% \end{wrapfigure}

\subsection{Communication Patterns}
\label{sec:appendix_a}
To collect the trajectories and the reward signal data from different types of tasks, we design the communication patterns for these tasks and unified the data format as described in \fig{fig:buffer}. Here we use three python-sytle algorithms (\Algref{alg:monologue} \Algref{alg:dialogue} \Algref{alg:analogue}) to demonstrate how three types of communication patterns help the agent collect exploration data. 
\begin{algorithm}[ht]

% \algcomment{\fontsize{7.2pt}{0em}\selectfont \texttt{bmm}: batch matrix multiplication; \texttt{mm}: matrix multiplication; \texttt{cat}: concatenation.
%\vspace{-1.em}
% }
\definecolor{codeblue}{rgb}{0.25,0.5,0.5}
\lstset{
  backgroundcolor=\color{white},
  basicstyle=\fontsize{7.2pt}{7.2pt}\ttfamily\selectfont,
  columns=fullflexible,
  breaklines=true,
  captionpos=b,
  commentstyle=\fontsize{7.2pt}{7.2pt}\color{codeblue},
  keywordstyle=\fontsize{7.2pt}{7.2pt},
%  frame=tb,
}
\begin{lstlisting}[language=python]
# agent1: LLaMA agent
# agent2: GPT-4 agent
# input:  Task description 
# output: S  = (T, M, R)

# initialization
T, M, R = [input], [0], [0]

i = 0
while i < max_steps:
    T += ["think:"]
    thought = agent2.api(T)
    T.append(thought)
    M.append(2) # teacher agent message mask
    R.append(0)
    

    T += ["act:"]
    action = agent1.api(T)
    T.append(action)
    M.append(1) # student agent message mask
    R.append(0)

    response = env.excute(action)
    reward = parse(response)
    T.append(response)
    M.append(0) # system message mask
    R.append(reward)
    
    i += 1
    if reward != 0:
        break
S  = (T, M, R)   
return S
\end{lstlisting}
\caption{The Python-style algorithm to demonstrate Dialogue pattern}
\label{alg:dialogue}
\end{algorithm}
% \end{minipage}
% \vspace{-14pt}
% \end{wrapfigure}

% \begin{wrapfigure}{R}{.5\textwidth}
% \begin{minipage}{.5\textwidth}
% % \vspace{-8mm}
\begin{algorithm}[ht]

% \algcomment{\fontsize{7.2pt}{0em}\selectfont \texttt{bmm}: batch matrix multiplication; \texttt{mm}: matrix multiplication; \texttt{cat}: concatenation.
%\vspace{-1.em}
% }
\definecolor{codeblue}{rgb}{0.25,0.5,0.5}
\lstset{
  backgroundcolor=\color{white},
  basicstyle=\fontsize{7.2pt}{7.2pt}\ttfamily\selectfont,
  columns=fullflexible,
  breaklines=true,
  captionpos=b,
  commentstyle=\fontsize{7.2pt}{7.2pt}\color{codeblue},
  keywordstyle=\fontsize{7.2pt}{7.2pt},
%  frame=tb,
}
\begin{lstlisting}[language=python]
# agent1: LLaMA agent
# agent2: GPT-4 agent
# input:  Question description 
# output: S  = (T, M, R)

# initialization
T, M, R = [input], [0], [0]

i = 0
while i < max_steps:
    T += ["answer the question step by step:"]
    answer1 = agent1.api(T)
    query = T + answer1 + ["the answer is correct, yes or no? also gives a better answer"]
    response = agent2.api(query)
    reward, answer2 = parse(response)
    T.append(answer1)
    T.append(answer2)
    M.append(1) # student agent message mask
    M.append(2) # teacher agent message mask
    R.append(reward)
    R.append(+1) # assume teacher is correct


    query = query + response + ["please generate a similar qa pair to teach the student:"]
    response = agent2.api(query)
    new_question, teacher_answer = parse(response)
    new_question += "answer the question step by step:"
    student_answer = agent1.api(new_question)
    reward = parse(student_answer, teacher_answer)
    T.append(new_question + student_answer)
    M.append(1) # student agent message mask
    R.append(reward) 

    i += 1
 
S  = (T, M, R)   
return S
\end{lstlisting}
\caption{The Python-style algorithm to demonstrate Analogue pattern}
\label{alg:analogue}
\end{algorithm}
% \end{minipage}
% \vspace{-14pt}
% \end{wrapfigure}

\subsection{Buffer Structure}
\label{sec:appendix_b}
% \myparagraph{Buffer Structure}
The communication data will be saved as replay buffers for the updating phase, and the buffer data format is a serial of tokens sequences demonstrated in \fig{fig:buffer}. We treat each token as the action unit in our reinforcement learning formula, and each exploration trail is processed into 5 data sequences $[\textbf{S}_\textbf{a}, \textbf{S}_\textbf{m}, \textbf{S}_\textbf{v}, \textbf{S}_\textbf{l}, \textbf{S}_\textbf{r}]$:

\begin{itemize}
  \item $\textbf{S}_\textbf{a}$: A list of integers representing the generated token ids encoded by the tokenizer. All the valid text trajectories are recorded as a queue, including system texts like environment descriptions, feedback, and agent texts like parsed actions, thinking processes, and hints from other agents. While the invalid generated text of the agent will be skipped, such as nonsense string and action text can not be parsed. These tokens are treated equally as the input for the LLM, but they have different masks to apply different losses.

  \item $\textbf{S}_\textbf{m}$: The system mask to mask different types of input tokens to control the training loss. We set 0 as the default mask for system texts like environment descriptions, system feedback, and system prompts, the actions encoded from these kinds of texts are not actions we want the agent to learn, so they will be masked out both policy loss and value loss in the PPO algorithm.  We set 1 as the mask for agents-generated tokens like the keywords of decisions and the thinking process, which are the main supervising objects of our reinforcement learning pipeline, so they will be assigned full policy loss and value loss. We set 2 as the mask for hints or feedback from other agents, which are the actions we also want our own agent to learn but without instant state values since they are not generated by our agent. So the tokens with mask 2 will be mask out only the value loss and supervised by the policy loss.  
  % \zc{ I feel the writing of this subsection is pretty tedious. You are  basically talking about all the implementation details. We need to change the way we write it. Make it concept-oriented, or innovation-oriented, instead of implementation-oriented. Instead of using those symbols as the headings, think about what are the new concepts or ideas you have designed for representing the  communication history and facilitating later learning . Use those concepts to anchor your writing. In this way, you can sell your new contributions, instead of making the readers feel tedious and confused in reading the details.}

  \item $\textbf{S}_\textbf{v}$: The state values corresponding to the actions obtained by the value head our the agent model. The value head is an addition layer inserted to the original pre-trained LLM architecture, we implement it by inserting a linear layer after the second-to-last LlamaDecoderLayer as the auxiliary output module and the output values are processed by a $tanh()$ function to keep it range inside $(-1, 1)$.
  \item $\textbf{S}_\textbf{r}$: The rewards corresponding to the actions. The rewards are very sparse, most of the actions are zero-reward, and only when the current task is finished or the token length of the current buffer has just overflowed it will be non-zero value: +1 for positive, -1 for negative.
\end{itemize}

\subsection{Algorithm of \modelshort}
\begin{wrapfigure}{R}{.5\textwidth}
\begin{minipage}{.5\textwidth}
% \vspace{-8mm}
\begin{algorithm}[H]

% \algcomment{\fontsize{7.2pt}{0em}\selectfont \texttt{bmm}: batch matrix multiplication; \texttt{mm}: matrix multiplication; \texttt{cat}: concatenation.
%\vspace{-1.em}
% }
\definecolor{codeblue}{rgb}{0.25,0.5,0.5}
\lstset{
  backgroundcolor=\color{white},
  basicstyle=\fontsize{7.2pt}{7.2pt}\ttfamily\selectfont,
  columns=fullflexible,
  breaklines=true,
  captionpos=b,
  commentstyle=\fontsize{7.2pt}{7.2pt}\color{codeblue},
  keywordstyle=\fontsize{7.2pt}{7.2pt},
%  frame=tb,
}
\begin{lstlisting}[language=python]
# agent: Pre-trained LLM agent
# n_gpu: total number of GPUs 
# env_cls: the class of environments
# n_gen: the generation size for one iteration
# n_train: the train size for one iteration

# initialization
agent = instruction_finetune(agent)
replay_buffer = []
i = 0
while i < max_iteration:
  i += 1
  # Exploration Phase
  envs = env_cls(sample(data, n_gen//n_gpu))
  # asynchronously generate
  new_buffer = generate_trials(agent, envs)
  # dist.gather and dist.broadcast
  new_buffer = sync_all_gpus(new_buffer) 
  replay_buffer.append(new_buffer)

  # Training Phase
  rollouts = sample(replay_buffer, n_train))
  # distributed training with ppo
  agent = ppo_ddp_train(agent, rollouts)
\end{lstlisting}
\caption{Python-style code of LTC}
\label{alg:code}
\end{algorithm}
\end{minipage}
\vspace{-14pt}
\end{wrapfigure}

The implementation of \modelshort can be summarized as \Algref{alg:code}, we unveil the structural framework that embodies the Learning Through Communication (LTC) paradigm, meticulously crafted to facilitate autonomous and progressive learning through iterative cycles. At the outset, the pre-trained Large Language Model (LLM) agent undergoes a fine-tuning phase to align its initial configuration with the intended learning tasks. Following this, an exploration phase is initiated where a batch of training environments is sampled and subsequently engaged by the agent to generate trial data asynchronously, leveraging the power of parallel computing across multiple GPUs to enhance efficiency. This newly generated data is synchronized across all GPUs to foster a cohesive learning base, which is stored in a replay buffer for further utilization. As the agent iterates through this process, it continually samples from this replay buffer during the updating phase, employing the Proximal Policy Optimization (PPO) algorithm in a distributed data parallel (DDP) setting to refine its strategies and adapt dynamically. This code snippet, therefore, encapsulates the innovative asynchronous and distributed nature of the LTC paradigm, marking a significant stride in fostering intelligent, adaptive, and collaborative artificial intelligence agents.

\begin{figure}[t]
  \centering
  \includegraphics[scale=0.32]{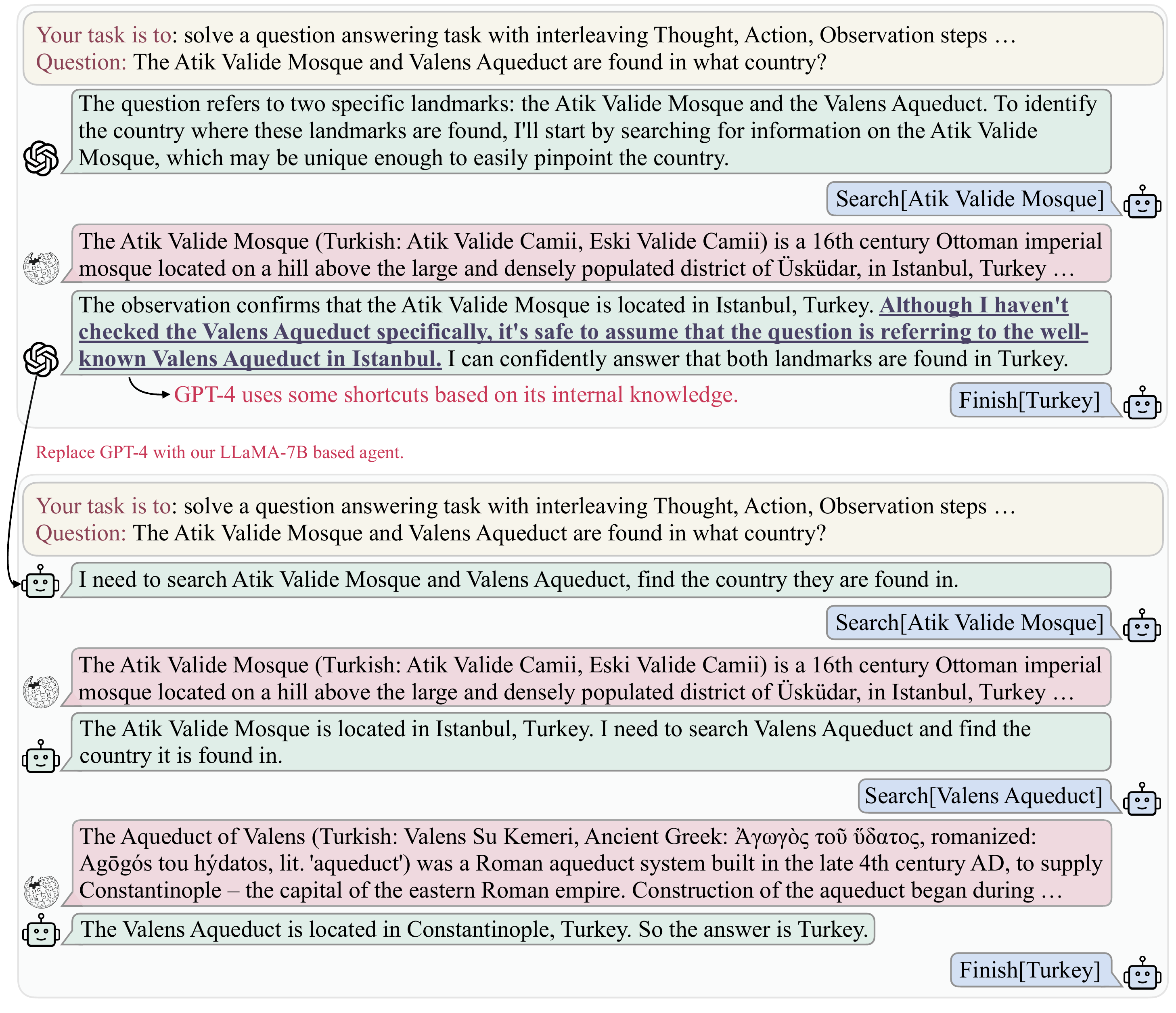}
  \caption{GPT-4 can use  shortcuts to solve the problem, while the LLaMA-7B agent cannot mimic it.}
  \label{fig:shortcut_compare}
\end{figure}
\subsection{Training Loss}

After exploration in each iteration, we update the replay buffer by incorporating the newly collected trajectories and then sample the most recent trajectories to 
train the parameters $\pi_\theta$ of the LLM agent. We design our training objective to combine: 1)
the standard language modeling loss $\mathcal{L}_{\text{LM}}$, 2) the policy loss $\mathcal{L}_{\text{policy}}$, 3) the value loss $\mathcal{L}_{\text{value}}$, and 4) the entropy loss $\mathcal{L}_{\text{entropy}}$.
 The  overall training objective is formulated as:
\begin{align*}
  \mathcal{L}_{\text{total}} &= \mathcal{L}_{\text{LM}} + \beta ( \mathcal{L}_{\text{policy}} + \lambda \mathcal{L}_{\text{value}} + \mathcal{L}_{\text{entropy}})
\end{align*}
where $\beta$ and $\lambda$ are weighting hyperparameters. 

The different losses in the above are described as follows:

\begin{itemize}
        \item

The LM loss $\mathcal{L}_{\text{LM}}$ is defined as the cross entropy between the agent and its generations which have a positive reward, akin to self-improving model schemes \cite{huang2022large, rafailov2023direct}. By training on these generations, the agent is further encouraged to give generations which yield positive rewards.

        \item
The policy loss $\mathcal{L}_{\text{policy}}$ is introduced to supervise the agent's actions. The policy loss $\mathcal{L}_{\text{policy}}$ is calculated using a masked version of the surrogate objective defined in \cite{schulman2017proximal} with advantage estimates $\hat{A}$, 
\[
\mathcal{L}_\text{policy}(\theta) = -\mathbb{E} [ m_{\text{policy}} * \text{min}(r(\theta)\hat{A}, \text{clip}(r(\theta), 1-\epsilon, 1+\epsilon)\hat{A}  ],
\]
where $r(\theta)$ is the output probability ratio $r(\theta) = \frac{  \pi_\theta(a\: |\: s)   }{  \pi_{\text{old}}(a \: | \: s)   }$ of the agent with its previous version $\pi_{\text{old}}$. We define binary mask $m_{\text{policy}}$ to mask out the encoded system message in PPO loss (marked by $S_m=0$ in buffers \ref{sec:appendix_b}). For example, let $\{x_1, y_1, x_2, y_2, \dots  x_n, y_n  \}$ be a token buffer consisting of system messages $x_n \in X$ and agents' messages (include the target trained agent and the other teacher agents) $\pi_\theta$ output $y_n \in Y$, then the binary mask $m_{\text{policy}} =  \{0, 1, 0, 1, \dots  0, 1  \}$. 

        \item
The value loss is defined in \cite{schulman2017proximal} as the mean squared error between calculated value and estimated advantages masked by another binary mask $m_{\text{value}}$ (marked by $S_m=1$ in buffers \ref{sec:appendix_b}). For example, let $\{z_1, y_1, z_2, y_2, \dots  z_n, y_n  \}$ be a token buffer consisting of all other messages (except the agent-generated messages) $z_n \in X$ and trained agent-generated messages $\pi_\theta$ output $y_n \in Y$, then the binary mask $m_{\text{policy}} =  \{0, 1, 0, 1, \dots  0, 1  \}$.

        \item
$\mathcal{L}_{\text{entropy}} $ is an entropy bonus to ensure sufficient exploration, as suggested in past work \citep{williams1992simple, mnih2016asynchronous}. This entropy is computed as a small negative factor times the entropy of the policy distribution :
\(\mathcal{L}_{\text{entropy}} = 0.01 \times \sum_{a} \pi_{\theta}(a|s) \log \pi_{\theta}(a|s) \). 

\end{itemize}

\subsection{Implementation detail}
\label{sec:appendix_c}
\subsection{Asynchronously Distributed Generating}
The exploration data is generated in an asynchronous style, so that the agent can handle the environments with open-end exploration space. The training data are pre-processed into interactive environments which are capable for agents to observe the states, take actions, and get immediate feedback. According to the number of GPU threads, these environments are divided into corresponding portions and then distributed to each GPU. Subsequently, these GPUs begin to explore these environments asynchronously in parallel with the same agent trained by the latest data. Since the lengths of the generated contents are varied and the interactions inside the environments are generally open-ended, the time cost for the agent to explore each environment is also varied, some GPU threads may process the data faster than others. A barrier is set for all the GPU threads so that the early finished GPU threads can wait for the others until the total accumulated buffers generated by the environments achieve a preset number $S_{g}$, which is the quantity of the new training buffers we want to add to the replay buffers in one iteration. After all the GPU threads reach the barrier, we get enough buffers then gather the buffers from each GPU thread and merge them together, and broadcast the new buffers to each GPU thread to update their local replay buffers. The updated replay buffers will be used in the updating phase for training the agents of the next iteration.

% \subsection{System Mask}
%   % \item $\textbf{S}_\textbf{m}$: 
%   The system mask to mask different types of input tokens to control the training loss. We set 0 as the default mask for system texts like environment descriptions, system feedback, and system prompts, the actions encoded from these kinds of texts are not actions we want the agent to learn, so they will be masked out both policy loss and value loss in the PPO algorithm.  We set 1 as the mask for agents-generated tokens like the keywords of decisions and the thinking process, which are the main supervising objects of our reinforcement learning pipeline, so they will be assigned full policy loss and value loss. We set 2 as the mask for hints or feedback from other agents, which are the actions we also want our own agent to learn but without instant state values since they are not generated by our agent. So the tokens with mask 2 will be mask out only the value loss and supervised by the policy loss. 

\subsection{Baselines}
\label{sec:appendix_d}
\textbf{ReAct}~\citep{yao2023react} uses a subset of training cases as prompts for different tasks, in the format of thought-action-observation sequences. For knowledge-intensive reasoning tasks like \textit{HotpotQA}, ReAct designs an action space that includes search, lookup, and finish actions, enabling the agent to interact with Wikipedia to retrieve necessary information.
On the other hand, \textbf{ReAct-IM} adopts Inner Monologue (IM)~\citep{huang2022inner} style prompting. Chain-of-thought prompting (\textbf{CoT})~\citep{wei2022chain}, enhances the reasoning capabilities of Language and Vision models (LLMs) by generating a sequence of intermediate reasoning steps. This can be considered as a reasoning-only baseline of ReAct, excluding actions and observations. Additionally, \textbf{CoT-SC}~\citep{wang2022self-consistency,wang2022rationale} is a follow-up work of CoT, serving as a self-consistency baseline. It is worth noting that most of these methods employ greedy decoding, except for BUTLER~\cite{micheli2021language}, which utilizes beam search. Most of these methods focus on few-shot prompting, and different pre-trained models are used. To ensure a fair comparison, we include the additional baselines named ReAct-Tuning and CoT-Tuning by fine-tuning the LLaMA-7B model using the collected trajectories as fine-tuning data mentioned in~\ref{par:pretraining}. In addition, GPT-4 are not used in the test time, and all the results reported are obtained by the trained agent itself.

\subsection{Losses}
We conducted ablation studies on the loss design of LTC. Figure \fig{fig:curves_ppo} illustrates the success rate of agents on the \textit{ALFWorld} dataset under different loss settings. Without using our communication pattern for interactions and merely sampling pre-collected instruction data for training, the improvement was limited. However, when we incorporated our communication pattern to gather data, the model's performance quickly surpassed 80\%. 
Furthermore, employing PPO loss to handle positive and negative samples separately resulted in faster and more significant improvement (blue line).
In Figure \fig{fig:curves_ppo_loss}, we present the separate curves of the three main losses during training. Initially, the LM loss showed a decreasing trend. Interestingly, as training iterations progressed, both the value loss and policy loss gradually decreased, which possibly causes the LM loss to increase temporarily. After the value loss and policy loss reached a certain threshold, the LM loss continued to decrease till convergence.

\begin{figure*}[t]
  \centering
  \includegraphics[scale=0.45]{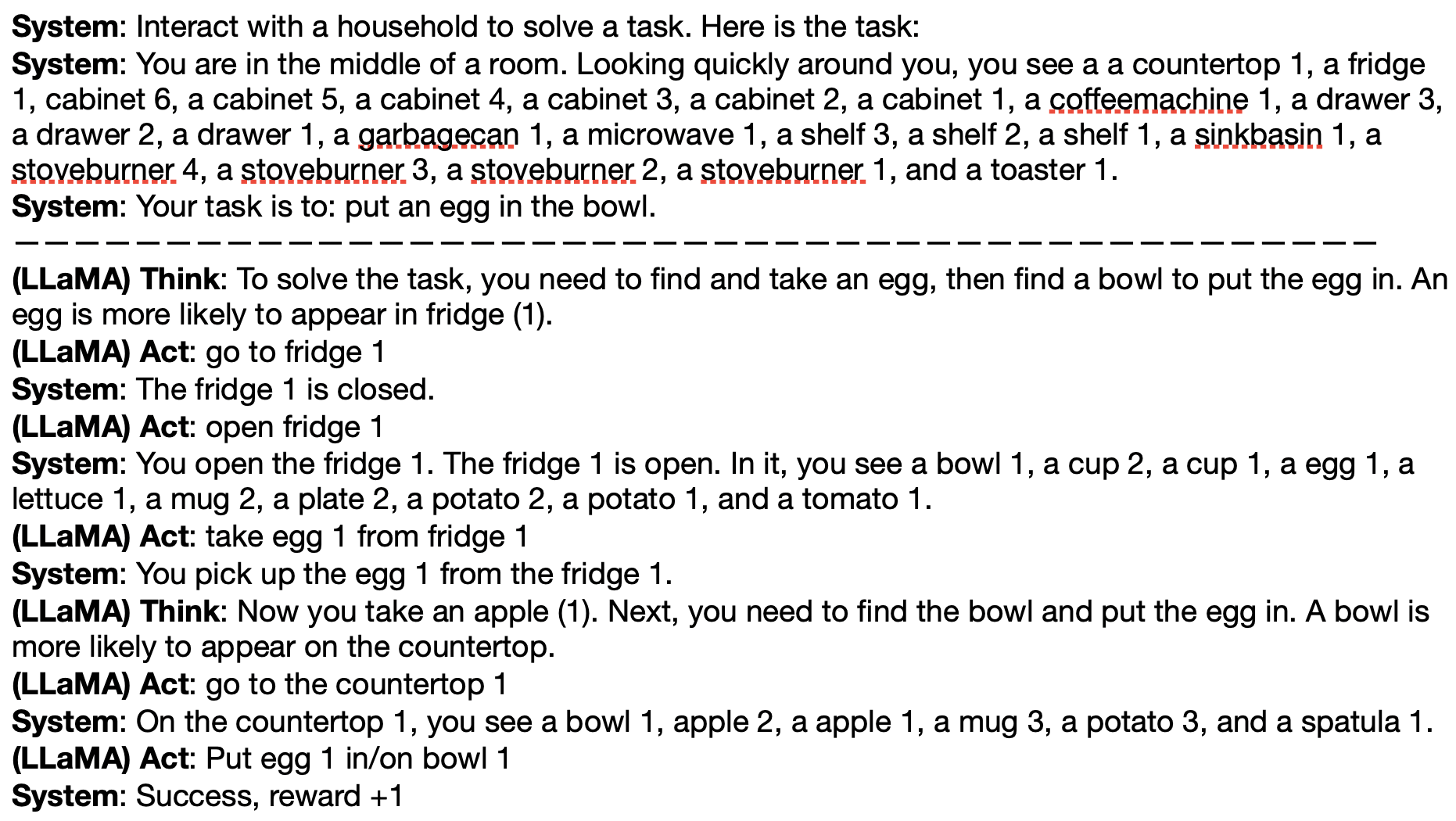}
  \caption{The text version of the toy example in ~\fig{fig:teaser} for Monologue pattern of ALFWorld. }
  \label{fig:teaser_text}
\end{figure*}
% \section{Cases study}
\subsection{Cases study}
One interesting observations is that the GPT-4 agent sometimes employs "shortcuts" to solve problems when serving as a teacher to generate new training data. These shortcuts rely on the internal knowledge acquired during its pretraining process. To illustrate this, we present a case study from \textit{HotpotQA} in Figure \fig{fig:shortcut_compare}. In this case, the GPT-4 agent quickly retrieves the answer by leveraging its memorized knowledge about the second entry after receiving the Wikipedia page of the first entry. On the other hand, the bottom of Figure \fig{fig:shortcut_compare} demonstrates a comparison with LLaMA-7B, which was trained using our LTC  method with the GPT-4 agent in the loop. LLaMA-7B does not employ shortcuts and instead performs a search for the second entry. This case study   demonstrates that communication mechanism in LTC provide additional benefits during learning, compared to soley relying on data generated by GPT-4.

%%%%%%%%%%%%%%%%%%%%%%%%%%%%%%%%%%%%%%%%%%%%%%%%%%%%%%%%%%%%%%%%%%%%%%%%%%%%%%%
%%%%%%%%%%%%%%%%%%%%%%%%%%%%%%%%%%%%%%%%%%%%%%%%%%%%%%%%%%%%%%%%%%%%%%%%%%%%%%%

\end{document}